%% file: template.tex
\documentclass{article}

\usepackage{arxiv}

\usepackage[utf8]{inputenc} % allow utf-8 input
\usepackage[T1]{fontenc}    % use 8-bit T1 fonts
\usepackage{hyperref}       % hyperlinks
\usepackage{url}            % simple URL typesetting
\usepackage{booktabs}       % professional-quality tables
\usepackage{amsfonts}       % blackboard math symbols
\usepackage{nicefrac}       % compact symbols for 1/2, etc.
\usepackage{microtype}      % microtypography
\usepackage{lipsum}
\usepackage{graphicx}
\usepackage{mdframed} %for framing
\usepackage{framed}
\usepackage{fancyhdr}
\usepackage[dvipsnames]{xcolor}
\usepackage[backend=biber,style=ieee, maxnames=10, sorting=none]{biblatex}
\bibliography{references} 

\usepackage{cleveref}
\crefname{section}{Sec.}{Sec.}
\crefname{Section}{Sec.}{Sec.}
\crefname{table}{Tab.}{Tab.}
\crefname{appendix_table}{Tab.}{Tab.}
\crefname{Table}{Tab.}{Tab.}
\crefname{Figure}{Fig.}{Fig.}
\crefname{figure}{Fig.}{Fig.}
\crefname{appendix}{Appendix}{Appendix}
\crefname{chapter}{Chapter}{Chapter}

\title{Auditing large language models: \\
a three-layered approach}

\author{
 Jakob M{\"o}kander$^{1,2,}$\thanks{Corresponding author: \texttt{jakob.mokander@oii.ox.ac.uk}}\\
   \And
 Jonas Schuett$^{3,4}$ \\
  \And
 Hannah Rose Kirk$^{1}$ \\
 \And
 Luciano Floridi$^{1,5}$
  \And
  \vspace{-1em}
  \\
  $^1$ Oxford Internet Institute, University of Oxford, Oxford, UK\\
  $^2$ Center for Information Technology Policy, Princeton University, Princeton, US \\
  $^3$ Centre for the Governance of AI, Oxford, UK \\
  $^4$ Faculty of Law, Goethe University Frankfurt, Frankfurt am Main, Germany \\
  $^5$ Department of Legal Studies, University of Bologna, Bologna, Italy
}

\date{\textcolor{RedViolet}{This paper is accepted for publication in AI \& Ethics. The official citation is: \\ M{\"o}kander, J., Schuett, J., Kirk, H.R., \& Floridi, L., Auditing large language models: \\ a three-layered approach. \textit{AI Ethics} (2023). \href{https://doi.org/10.1007/s43681-023-00289-2}{\texttt{doi:10.1007/s43681-023-00289-2}}}\\}

\begin{document}
\maketitle
\begin{abstract}
Large language models (LLMs) represent a major advance in artificial intelligence (AI) research. However, the widespread use of LLMs is also coupled with significant ethical and social challenges. Previous research has pointed towards auditing as a promising governance mechanism to help ensure that AI systems are designed and deployed in ways that are ethical, legal, and technically robust. However, existing auditing procedures fail to address the governance challenges posed by LLMs, which display emergent capabilities and are adaptable to a wide range of downstream tasks. In this article, we address that gap by outlining a novel blueprint for how to audit LLMs. Specifically, we propose a three-layered approach, whereby governance audits (of technology providers that design and disseminate LLMs), model audits (of LLMs after pre-training but prior to their release), and application audits (of applications based on LLMs) complement and inform each other. We show how audits, when conducted in a structured and coordinated manner on all three levels, can be a feasible and effective mechanism for identifying and managing some of the ethical and social risks posed by LLMs. However, it is important to remain realistic about what auditing can reasonably be expected to achieve. Therefore, we discuss the limitations not only of our three-layered approach but also of the prospect of auditing LLMs at all. Ultimately, this article seeks to expand the methodological toolkit available to technology providers and policymakers who wish to analyse and evaluate LLMs from technical, ethical, and legal perspectives.
\end{abstract}

\keywords{artificial intelligence \and auditing \and ethics \and foundation models \and governance \and large language models \and \\natural language processing \and policy \and risk management}

%% SECTIONS
\setcounter{footnote}{0}
\input{sections/1_introduction.tex}
\input{sections/2_need_to_audit_LLMs.tex}

\input{sections/3_merits_and_limits.tex}

\input{sections/4_auditing_LLMs.tex}

\input{sections/5_limitations.tex}
\input{sections/6_conclusion.tex}

\paragraph{Authorship statement} JM is the first author of this paper. JS, HRK, and LF contributed equally to the paper.

\paragraph{Acknowledgements} The authors would like to thank the following people for helpful comments on earlier versions of this manuscript: Markus Anderljung, Matthew Salganik, Arvind Narayanan, Deep Ganguli, Katherine Lee, Toby Shevlane, Varun Rao, and Lennart Heim. In the process of writing the article, the authors also benefitted from 
conversations with and input from Allan Dafoe, Ben Garfnkel, Anders Sandberg, Emma Bluemke, Andreas Hauschke, Owen Larter, Sebastien Krier, Mihir Kshirsagar, and Jade Leung. The article is much better for it.

\paragraph{Conflicts of interest} The authors have no conflicts of interest to declare.

\paragraph{Funding} JM's doctoral research at the Oxford Internet Institute is supported through a studentship provided by AstraZeneca. JM conducted part of this research during a paid Summer Fellowship at the Centre for Governance of AI. HRK's doctoral research at the Oxford Internet Institute is supported by the UK Economic and Social Research Council grant ES/P000649/1.

\input{sections/7_appendix.tex}

\newpage
\printbibliography
% \bibliographystyle{unsrt}
% \bibliography{references}  %%% Remove comment to use the external .bib file (using bibtex).

% %%% Comment out this section when you \bibliography{references} is enabled.

% George Kour and Raid Saabne.
% \newblock Real-time segmentation of on-line handwritten arabic script.
% \newblock In {\em Frontiers in Handwriting Recognition (ICFHR), 2014 14th
%   International Conference on}, pages 417--422. IEEE, 2014.

% George Kour and Raid Saabne.
% \newblock Fast classification of handwritten on-line arabic characters.
% \newblock In {\em Soft Computing and Pattern Recognition (SoCPaR), 2014 6th
%   International Conference of}, pages 312--318. IEEE, 2014.

% \bibitem{hadash2018estimate}
% Guy Hadash, Einat Kermany, Boaz Carmeli, Ofer Lavi, George Kour, and Alon
% \newblock Estimate and replace: A novel approach to integrating deep neural
%   networks with existing applications.
% \newblock {\em arXiv preprint arXiv:1804.09028}, 2018.

% \end{thebibliography}

\end{document}

%% file: sections/1_introduction.tex
\section{Introduction}
Auditing is a governance mechanism that technology providers and law enforcers can use to identify and mitigate risks associated with artificial intelligence (AI) systems \cite{sandvigAuditing2014, diakopoulosAlgorithmic2015,mokanderEthics2021,brundageTrustworthy2020a, rajiActionable2019}.\footnote{The term \textit{governance mechanism} refers to the set of activities and controls used by various parties in society to exert influence and achieve normative ends \cite{Baldwin2011}.} Specifically, auditing is a systematic and independent process of obtaining and evaluating evidence regarding an entity's actions or properties and communicating the results of that evaluation to relevant stakeholders \cite{mokanderEthicsbased2021}. Three ideas underpin the promise of auditing as an AI governance mechanism: that procedural regularity and transparency contribute to good governance \cite{cobbeReviewable2021, floridiInfraethics2017}; that proactivity in the design of AI systems helps identify risks and prevent harm before it occurs \cite{rajiClosing2020, koshiyamaAlgorithm2022}; and, that the independence between the auditor and the auditee contributes to the objectivity and professionalism of the evaluation \cite{powerAudit1997, rajiOutsider2022}.

Previous work on AI auditing has focused on ensuring that specific applications meet predefined, often sector-specific, requirements. For example, researchers have developed procedures for how to audit AI systems used in recruitment \cite{kazimSystematizing2021}, online search \cite{robertsonAuditing2018} and medical diagnostics \cite{oakden-raynerValidation2022, liuMedical2022}. However, the capabilities of AI systems tend to become ever more general. In a recent article, Bommasani et al. \cite{bommasaniOpportunities2022} coined the term \textit{foundation models} to describe models that can be adapted to a wide range of downstream tasks via transfer learning. While foundation models are not necessarily new from a technical perspective, they differ from other AI systems insofar as they have proven to be effective across many different tasks and display emergent capabilities when scaled \cite{bommasaniReflections2021}.\footnote{Foundation models are typically based on deep neural networks and self-supervised learning, two approaches that have existed for decades \cite{bommasaniOpportunities2022}. That said, the rise of foundation models has been enabled by more recent developments, including: the advancement of new network architectures, like transformers \cite{Vaswani2017}; the increase in compute resources and improvements in hardware capacity \cite{SmithGoodson2022}; the availability of large scale datasets, e.g., through ImageNet \cite{Russakovsky2015} or CommonCrawl \cite{Luccioni2021}; and the application of these increased compute resources with larger datasets for model pre-training \cite{Han2021}.} The rise of foundation models also reflects a shift in how AI systems are designed and deployed, since these models are typically trained and released by one actor and subsequently adapted for a plurality of applications by other actors.

From an AI auditing perspective, foundation models pose significant challenges. For example, it is difficult to assess the risks that AI systems pose independent of the context in which they are deployed. Moreover, how to allocate responsibility between technology providers and downstream developers when harms occur remains unresolved. Taken together, the capabilities and training processes of foundation models have outpaced the development of tools and procedures to ensure that these are ethical, legal, and technically robust.\footnote{The European Commission's \textit{Ethics Guidelines for Trustworthy AI} stipulate that AI systems should be legal, ethical, and technically robust \cite{commissionEthics2019}. That normative standard includes safeguards against both immediate and long-term concerns, e.g., those related to data privacy and wrongful discrimination and those related to the safety and control of AI systems, respectively.}  This implies that, while application-level audits have an important role in AI governance, they must be complemented with new forms of supervision and control.

This article addresses that gap by focusing on a subset of foundation models, namely large language models (LLMs). Language models start from a source input, called the prompt, to generate the most likely sequences of words, code, or other data \cite{Floridi2020}. Historically, different model architectures have been used in natural language processing (NLP), including probabilistic methods \cite{rosenfeldTwo2000}. However, most recent LLMs – including those we focus on in this article – are based on deep neural networks trained on a large corpus of texts. Examples of such LLMs include GPT-3 \cite{brownLanguage2020}, PaLM \cite{chowdheryPaLM2022a}, LaMDA \cite{thoppilanLaMDA2022a}, Gopher \cite{raeScaling2022a} and OPT \cite{zhangOPT2022}. Once an LLM has been pre-trained, it can be adapted (with or without fine-tuning\footnote{To fine-tune LLMs for specific tasks, an additional dataset of in-domain examples can be used to adapt the final layers of a pre-trained model. In some cases, developers apply reinforcement learning (RL) – a feedback driven training paradigm whereby LLMs learn to adjust their behaviour to maximise a reward function \cite{Korbak2022}; especially reinforcement learning from human feedback (RLHF) – where the reward function is estimated based on human ratings of model outputs \cite{Bai2022, Stiennon2020}. LLMs can also be adapted with no additional training data and frozen weights – via in-context learning or prompt-based demonstrations \cite{Min2022}.}) to support various applications, from spell-checking \cite{huMisspelling2021a} to creative writing \cite{hsiehTransformer2019}.

Developing LLM auditing procedures is a relevant task for two reasons. First, previous research has demonstrated that LLMs pose many ethical and social challenges, including the perpetuation of harmful stereotypes, the leakage of personal data protected by privacy regulations, the spread of misinformation, plagiarism, and the misuse of copyrighted material \cite{weidingerEthical2021b, benderDangers2021, shelbyIdentifying2023}. In recent months, the scope of impact from these harms has been dramatically scaled by unprecedented public visibility and growing user bases of LLMs. For example, ChatGPT attracted over 100 million users just two months after its launch \cite{Curry2023}. The urgency of addressing those challenges makes developing a capacity to audit LLMs' characteristics along different normative dimensions (such as privacy, bias, IP, etc.) a critical task in and of itself \cite{liangHolistic2022a}. Second, LLMs can be considered proxies for other foundation models.\footnote{In some cases, the ability to utilize other modalities is integrated into single-modal LLMs, e.g., DeepMind's Flamingo model \cite{alayracFlamingo2022a} fuses an LLM with visual embeddings to exploit its strong existing performance on text-based tasks.} For example, CLIP \cite{radfordLearning2021} is a vision-language model trained to predict which text caption accompanied an image. Although not an LLM, CLIP – and other models that can be adapted for multiple downstream applications – faces similar governance challenges. The same holds of other automatic content producers, such as DALL·E 2 \cite{rameshHierarchical2022}. So, developing LLM auditing procedures may inform the auditing of other foundation models and even more powerful generative systems in the future.\footnote{Following Jonathan Zittrain \cite{zittrainGenerative2014}, we define `generative technologies' as technologies that allow third parties to innovate upon them without any gatekeeping. Colloquially, `generative AI' sometimes refers to systems that can output content (images, text, audio, or code) \cite{openaiGenerative2016}, but that is not how we use the term in this article.} 

The main contribution offered in this article is a novel blueprint for how to audit LLMs. Specifically, we propose a three-layered approach, whereby \textit{governance audits} (of technology providers that design and disseminate LLMs), \textit{model audits} (of LLMs after pre-training but prior to their release), and \textit{application audits} (of applications based on LLMs) complement and inform each other. \cref{fig:1} provides an overview of this three-layered approach. As we demonstrate throughout this article, many tools and methods already exist to conduct audits at each individual level. However, the key message we seek to stress is that, to provide meaningful assurance for LLMs, audits conducted on the governance, model, and application levels must be combined into a structured and coordinated procedure. \cref{fig:2} illustrates how outputs from audits on one level become inputs for which audits on other levels must account. To our best knowledge, our blueprint for how to audit LLMs is the first of its kind, and we hope it will inform both technology providers’ and policymakers’ efforts to ensure that LLMs are legal, ethical, and technically robust.

In the process of introducing and discussing our three-layered approach, the article also offers two secondary contributions. First, it makes seven claims about how LLM auditing procedures should be designed to be feasible and effective in practice. Second, it identifies the conceptual, technical, and practical limitations associated with auditing LLMs. Together, these secondary contributions lay a groundwork that other researchers and practitioners can build upon when designing new, more refined, LLM auditing procedures in the future.

Our efforts tie into an extensive research agenda and policy formation process. AI labs like Cohere, OpenAI, and AI21 have expressed interest in understanding what it means to develop LLMs responsibly \cite{openaiBest2022}, and DeepMind, Microsoft, and Anthropic have highlighted the need for new governance mechanisms to address the social and ethical challenges that LLMs pose \cite{weidingerEthical2021b, peyrardInvariant2022, ganguliPredictability2022}. Individual parts of our proposal (e.g., those related to model evaluation \cite{chowdheryPaLM2022a} and red teaming \cite{Ganguli2022, perezRed2022a}) have thus already started to be implemented across the industry, although not always in a structured manner or with full transparency. Policymakers, too, are interested in ensuring that societies benefit from LLMs while managing the associated risks. Recent examples of proposed AI regulations include the EU AI Act \cite{commissionArtificial2021} and the US Algorithmic Accountability Act of 2022 \cite{wydenAlgorithmic2022}. The blueprint for auditing LLMs outlined in this article neither seeks to replace existing best practices for training and testing LLMs nor to foreclose forthcoming AI regulations. Instead, it complements them by demonstrating how governance, model, and application audits – when conducted in a structured and coordinated manner – can help ensure that LLMs are designed and deployed in ethical, legal, and technically robust ways.

A further remark is needed to narrow down this article’s scope. Our three-layered approach concerns the procedure of LLM audits and answers questions about what should be audited, when, and according to which criteria. Of course, when designing a holistic auditing ecosystem, several additional considerations exist, e.g., who should conduct the audit and how to ensure post-audit action \cite{rajiOutsider2022}. While such considerations are important, they fall outside the scope of this article. How to design an institutional ecosystem to audit LLMs is a non-trivial question that we have neither the space nor the capacity to address here. That said, the policy process required to establish an LLM auditing ecosystem will likely be gradual and involve negotiations between numerous actors, including AI labs, policymakers, and civil rights groups. For this reason, our early blueprint for how to audit LLMs is intentionally limited in scope to not forego but rather to initiate this policy formation process by eliciting stakeholder reactions.

The remainder of this article proceeds as follows. \cref{sec:2} highlights the ethical and social risks posed by LLMs and establishes the need to audit them. In doing so, it situates our work in relation to recent technological and societal developments. \cref{sec:3} reviews previous literature on AI auditing to identify transferable best practices, discusses the properties of LLMs that undermine existing AI auditing procedures, and derives seven claims for how LLM auditing procedures should be designed to be feasible and effective. \cref{sec:4} outlines our blueprint for how to audit LLMs, introducing a three-layered approach that combines governance, model, and application audits. The section explains in detail why these three types of audits are needed, what they entail, and the outputs they should produce. \cref{sec:5} discusses the limitations of our three-layered approach and demonstrates that any attempt to audit LLMs will face several conceptual, technical, and practical constraints. Finally, \cref{sec:6} concludes by discussing the implications of our findings for technology providers, policymakers, and independent auditors.

%% file: sections/2_need_to_audit_LLMs.tex
\section{The need to audit LLMs}
\label{sec:2}
This section summarises previous research on LLMs and their ethical and social challenges. It aims to situate our work in relation to that research, stress the need for auditing procedures that capture the risks LLMs pose, and address potential objections to our approach.

\subsection{The opportunities and risks of LLMs}
Although LLMs represent a major advance in AI research, the idea of building text-processing machines is not new. Since the 1950s, NLP researchers and practitioners have been developing software that can analyse, manipulate, and generate natural language \cite{joshiNatural1991}. Until the 1980s, most NLP systems used logic-based rules and focused on automating the structural analysis of language needed to enable machine translation and speech recognition \cite{hirschbergAdvances2015}. More recently, the advent of deep learning, advances in neural architectures such as transformers, growth in computational power and the availability of internet-scraped training data have revolutionised the field \cite{chernyavskiyTransformers2021} by permitting the creation of LLMs that can approximate human performance on some benchmarks \cite{adiwardanaHumanlike2020b,wangSuperGLUE2019}. Further advances in instruction-tuning and reinforcement learning from human feedback have improved model capabilities to predict user intent and respond to natural language requests \cite{Stiennon2020, Ouyang2022, arcasLarge2022}. 

LLMs' core training task is to produce the most likely continuation of a text sequence \cite{arcasLarge2022}. Consequently, LLMs can be used to recognise, summarise, translate, and generate texts, with near human-like performance on some tasks \cite{Suzgun2022}. Exactly when a language model becomes ``large'' is a matter of debate – referring to either more trainable parameters \cite{villalobosMachine2022}, a larger training corpus or a combination of these \cite{hoffmannTraining2022}. For our purposes, it is sufficient to establish that LLMs provide three key opportunities. First, they are highly adaptable to various downstream applications, requiring fewer in-domain labelled examples than traditional deep learning systems. LLMs can act as intermediary assets that are adapted for specific tasks, such as diagnosing medical conditions \cite{rasmyMedBERT2021}, generating code \cite{wangCodeT52021, chenEvaluating2021a} and translating languages \cite{wangLanguage2021a}. In some cases, LLMs can perform highly on a task with few-shot or zero-shot reasoning \cite{brownLanguage2020, kojimaLarge2023}.\footnote{An LLM is considered a `zero-shot' reasoner if employed for a completely unseen task and a `few-shot' reasoner if only a small sample of demonstrations are given for a previously unseen task.} Second, a scaling law has been identified whereby the training error of an LLM falls off as a power of training set size, model size or both \cite{kaplanScaling2020b}. Furthermore, simply scaling the model can result in emergent gains on a wide array of tasks \cite{srivastavaImitation2022a}, though those gains are non-uniform, especially for complex mathematical or logical reasoning domains \cite{raeScaling2022a}. Third, while some pre-trained models are protected by paywalls or siloed within companies, many LLMs are accessible via open-source libraries such as HuggingFace, democratising the gains from deep language modelling and allowing non-experts to use it in their applications \cite{kirkBias2021}.

Alongside such opportunities, however, the use of LLMs is coupled with ethical challenges \cite{benderDangers2021, shelbyIdentifying2023}. As recent controversies surrounding ChatGPT \cite{Azaria2022} have shown, LLMs are prone to give biased or incorrect answers to user queries \cite{Borji2023}. More generally, a recent article by Weidinger et al. \cite{weidingerEthical2021b} suggests that the risks associated with LLM include the following:
\begin{enumerate}
    \item \textit{Discrimination}: LLMs can introduce representational and allocational harms by perpetuating social stereotypes and biases;
    \item \textit{Information hazards}: LLMs may compromise privacy by leaking private information and inferring sensitive information;
    \item \textit{Misinformation hazards}: LLMs producing misleading information can lead to less well-informed users and erode trust in shared information;
    \item \textit{Malicious use}: LLMs can be co-opted by users with bad intent, e.g., to generate personalised scams or large-scale fraud;
    \item \textit{Human-computer interaction harms}: users may overestimate the capabilities of LLMs that appear human-like and use them in unsafe ways; and
    \item \textit{Automation and environmental harms}: training and operating LLMs require lots of computing power, incurring high environmental costs.
\end{enumerate}

Each of these risk areas constitutes a vast and complex field of research. Providing a comprehensive overview of each field's nuances is beyond this paper's scope. Instead, we take Weidinger et al.'s summary of the ethical and social risks associated with LLMs as a starting point for pragmatic problem-solving.

\subsection{The governance gap}
From a governance perspective, LLMs pose both methodological and normative challenges. As previously mentioned, foundation models – like LLMs – are typically developed and adopted in two stages. Firstly, a model is pre-trained using self-supervised learning on a large, unstructured text corpus scraped from the internet. Pre-training captures the general language representations required for many tasks without explicitly labelled data. Secondly, the weights or behaviours of this pre-trained model can be adapted on a far smaller dataset of labelled, task-specific, examples.\footnote{The term `adapted' here encompasses multiple existing methods for eliciting specific model behaviours, including fine-tuning, reinforcement learning with human feedback and in-context learning.} That makes it methodologically difficult to assess LLMs independent of the context in which they will be deployed \cite{bommasaniOpportunities2022}. 

Furthermore, although performance is predictable at a general level, performance on specific tasks, or at scale, can be unpredictable \cite{ganguliPredictability2022}. Crucially, even well-functioning LLMs force AI labs and policymakers to face hard questions, such as who should have access to these technologies and for which purposes \cite{shevlaneStructured2022}. Of course, the challenges posed by LLMs are not necessarily distinct from those associated with classical NLP or other deep-learning-based systems. However, LLMs' widespread use and generality make those challenges deserving of urgent attention. For all these reasons, analysing LLMs from ethical perspectives requires innovation in risk assessment tools, benchmarks, and frameworks \cite{tamkinUnderstanding2021a}.

Several governance mechanisms designed to ensure that LLMs are legal, ethical, and safe have been proposed or piloted \cite{avinFilling2021}. Some are technically oriented, including the pre-processing of the initial training data, the fine-tuning of LLMs on data with desired properties, and procedures to test the model at scale pre-deployment \cite{tamkinUnderstanding2021a, perezRed2022a}. Others seek to address the ethical and social risks associated with LLMs through sociotechnical mitigation strategies, e.g., creating more diverse developer teams \cite{paiResearching2020}, human-in-the-loop protocols \cite{wangPutting2021} and qualitative evaluation tools based on ethnographic methods \cite{mardaImportance2021}. Yet others seek to ensure transparency in AI development processes, e.g., through a structured use of model cards \cite{mitchellModel2019}, datasheets \cite{gebruDatasheets2021}, system cards \cite{metaaiSystem2023}, and the watermarking of system outputs \cite{Kirchenbauer2023}.\footnote{A watermark is a hidden pattern in a text that is imperceptible to humans but makes it algorithmically identifiable as synthetic.}

To summarise, while LLMs have shown impressive performance across a wide range of tasks, they also pose significant ethical and social risks. Therefore, the question of how LLMs should be governed has attracted much attention, with proposals ranging from structured access protocols designed to prevent malicious use \cite{shevlaneStructured2022} to hard regulation prohibiting the deployment of LLMs for specific purposes \cite{Hacker2023}. However, the effectiveness and feasibility of these governance mechanisms have yet to be substantiated by empirical research. Moreover, given the multiplicity and complexity of the ethical and social risks associated with LLMs, we anticipate that policy responses will need to be multifaceted and incorporate several complementary governance mechanisms. As of now, technology providers and policymakers have only started experimenting with different governance mechanisms, and how LLMs should be governed remains an open question \cite{Engler2023}.

\subsection{Call for Audits}
Against the backdrop of the technological and regulatory landscape surveyed in this section, auditing should be understood as one of several governance mechanisms different stakeholders can employ to ensure and demonstrate that LLMs are legal, ethical, and technically robust. It is important to stress that auditing LLMs is not a hypothetical idea but a tangible policy option that has been proposed by researchers, technology providers, and policymakers alike. For instance, when coining the term foundation models, Bommasani et al. \cite{bommasaniOpportunities2022} also suggested that ‘such models should be subject to rigorous testing and auditing procedures’. Moreover, in an open letter concerning the risks associated with LLMs and other foundation models, OpenAI’s CEO Sam Altman stated that ‘it’s important that efforts like ours submit to independent audits before releasing new systems’ \cite{Altman2023}. Finally, the European Commission is considering classifying LLMs and other foundation models as ‘high-risk AI systems’ \cite{Helberger2023}.\footnote{It is still uncertain how the EU AIA should be interpreted with respect to LLMs. The current formulation states that models that may be used for high-risk applications should be considered high-risk \cite{Bertuzzi2023}.} This would imply that technology providers designing LLMs have to undergo ‘conformity assessments with the involvement of an independent third-party’, i.e., audits by another name \cite{mokanderConformity2022}.

Despite widespread calls for LLM auditing, central questions concerning how LLMs can and should be audited have yet to be explored by academic researchers. This article addresses that gap by outlining a procedure for auditing LLMs. The main argument we advance can be summarised as follows. What auditing means varies between different academic disciplines and industry contexts \cite{Lee2008}. However, three strands of auditing research and practice are particularly relevant with respect to ensuring good governance of LLMs. The first stems from IT audits, whereby auditors assess the adequacy of technology providers’ software development processes and quality management procedures \cite{senftInformation2009}. The second strand stems from model testing and verification within the computer sciences, whereby auditors assess the properties of different computational models \cite{Dai2019}. The third strand stems from product certification procedures, whereby auditors test consumer goods for legal compliance and technical safety before they go to market \cite{Voas2016}. As we argue throughout this paper, it is necessary to combine auditing tools and procedural best practices from each of these three strands to identify and manage the social and ethical risks LLMs pose. Therefore, our blueprint for auditing LLMs combines governance audits of technology providers, model audits of LLMs, and application audits of downstream products and services built on top of LLMs. The details of this ‘three-layered approach’ are outlined in \cref{sec:4}.

\subsection{Addressing initial objections}
Before proceeding any further, it is useful to consider some reasonable objections to the prospect of auditing LLMs - as well as potential responses to these objections. First, one may argue that there is no need to audit LLMs per se and that auditing procedures should be established at the application level instead. Although audits on the application level are important, the objection presents a false dichotomy: quality and accountability mechanisms can and should be established at different stages of supply chains. Moreover, while some risks can only be addressed at the application level, others are best managed upstream. It is true that many factors, including some beyond the technology provider's control, determine whether a specific technological artefact causes harm \cite{deanAxes2021}. However, technology providers are still responsible for taking proportional precautions regarding reasonably foreseeable risks during the product life cycle stages that they do control. For this reason, we propose that application audits should be complemented with governance audits of the organisations that develop LLMs. The same logic underpins the EU's AI liability directive \cite{commissionAI2022}. Our proposal is thereby compatible with the emerging European AI regulations.

Second, identifying and mitigating all LLM-related risks at the technology level may not be possible. As we explain in \cref{sec:5}, this is partly because different normative values may conflict and require trade-offs \cite{berlinPursuit1988,gabrielArtificial2020,goodmanHard2021}. Using individuals’ data, for example, may permit improved personalisation of language models, but compromise privacy \cite{Kirk2023}. Moreover, concepts like ‘fairness’ or ‘transparency’ hide deep normative disagreements \cite{mittelstadtPrinciples2019}. For example, different definitions of fairness (like demographic parity and counterfactual fairness) are mutually exclusive \cite{mehrabiSurvey2021, kleinbergInherent2018, kusnerCounterfactual2017}, and prioritising between competing definitions remains a normative question. However, while audits cannot ensure that LLMs are ‘ethical’ in any universal sense, they nevertheless contribute to good governance in several ways. For example, audits can help technology providers identify risks and potentially prevent harm, shape the continuous (re-design) of LLMs, and inform public discourse concerning tech policy. Bringing all this together, our blueprint for how to audit LLMs focuses on making implicit choices and tensions visible, giving voice to different stakeholders, and generating resolutions that – even when imperfect – are, at least, more explicit and publicly defensible \cite{whittlestoneRole2019}.

Third, one may contend that designing LLM auditing procedures is difficult. We agree and would add that this difficulty has both practical and conceptual components. Different stages in the software development life cycle (including curating training data and the pre-training/fine-tuning of model weights) overlap in messy and iterative ways \cite{gururanganDon2020}. For example, consider open-source LLMs that are continuously re-trained and re-uploaded on collaborative platforms (like HuggingFace) post-release. That creates practical problems concerning when and where audits should be mandated. The conceptual challenges run even more deeply. For instance, what constitutes disinformation and hate speech are contested questions \cite{oneillPhilosopher2021}. Despite widespread agreement that LLMs should be `truthful' and `fair', such notions are hard to operationalise. Because there exists no universal condition of validity that applies equally to all kinds of utterances \cite{kasirzadehConversation2022c}, it is hard to establish a normative baseline against which LLMs can be audited.

However, these difficulties are not reasons for abstaining from developing LLM auditing procedures. Instead, they are healthy reminders that it cannot be assumed that one single auditing procedure will capture all LLM-related ethical risks or be equally effective in all contexts \cite{Steed2022}. The insufficiency and limited nature of auditing as a governance mechanism is not an argument against its complementary usefulness. With those caveats highlighted, we now review previous work on AI auditing. The aim of the next section is thus to explore the merits and limitations of existing AI.

%% file: sections/3_merits_and_limits.tex
\section{The merits and limits of existing AI auditing procedures}
\label{sec:3}
In this section, we review existing AI auditing procedures.\footnote{See \cref{sec:appendix} for the methodology used to conduct this literature review.} In doing so, we introduce auditing as an AI governance mechanism, highlight the properties of LLMs that undermine the feasibility and effectiveness of existing AI auditing procedures, and derive and defend seven claims about how LLM auditing procedures should be designed. Taken together, this section provides the theoretical justification for the LLM auditing blueprint outlined in \cref{sec:4}.

\subsection{AI Auditing}
In the broadest sense, auditing refers to an independent examination of any entity, conducted with a view to express an opinion thereon \cite{Gupta2004}. Auditing can be conceived as a governance mechanism because it can be used to monitor conduct and performance \cite{Flint1988} and has a long history of promoting procedural regularity and transparency in areas like financial accounting and worker safety \cite{LaBrie2019TowardsAF}. The idea behind AI auditing is thus simple: just like financial transactions can be audited for correctness, completeness, and legality, so can the design and use of AI systems be audited for technical robustness, legal compliance, or adherence with pre-defined ethics principles.

AI auditing is a relatively recent field of study, sparked in 2014 by Sandvig et al.’s article Auditing Algorithms \cite{sandvigAuditing2014}. However, auditing intersects with almost every aspect of AI governance, from the documentation of design procedures to model testing and verification \cite{Stodt2021}. AI auditing is thus both a multifaceted practice and a multidisciplinary field of research, harbouring contributions from computer science and engineering \cite{adler2018auditing, kearns2018preventing}, law \cite{Laux2021, selbstInstitutional2021}, media and communication studies \cite{sandvigAuditing2014, bandy2021problematic}, and organisation studies \cite{floridiCapAI2022, minkkinen2022continuous}.

Different researchers have defined AI auditing in different ways. For example, it is possible to distinguish between narrow and broad conceptions of AI auditing. The former is impact-oriented and focuses on probing and assessing the outputs of AI systems for different input data \cite{metaxaAuditing2021}. The latter is process-oriented and focuses on assessing the adequacy of technology providers’ software development processes and quality management systems \cite{berghoutAdvanced2023}. This article takes the broad perspective, defining AI auditing as a systematic and independent process of obtaining and evaluating evidence regarding an entity's actions or properties and communicating the results of that evaluation to relevant stakeholders. Note that the entity in question, i.e., the audit’s subject, can be either an AI system, an organisation, a process, or any combination thereof \cite{mokanderEthics2021a}.

Different actors can employ AI auditing for different purposes \cite{brownAlgorithm2021}. In some cases, policymakers mandate audits to ensure that AI systems used within their jurisdiction meet specific legal standards. For example, New York City’s AI Audit Law (NYC Local Law 144) requires independent auditing of companies utilising AI systems to inform employment-related decisions \cite{Dunn2023}. In other cases, technology providers commission AI audits to mitigate technology-related risks, calling on professional services firms like PwC, Deloitte, KPMG, and EY \cite{PwC2020, Deloitte2020, KPMG2020, EY2018}. In yet other cases, other stakeholders conduct AI audits to inform citizens about the conduct of specific companies.\footnote{This includes non-profit organisations like ForHumanity \cite{ForHumanity2023} and the Algorithmic Justice League \cite{AJL2023}.}

The key takeaway from this brief overview is that while AI auditing is already a widespread practice, both the design and purpose of different AI auditing procedures vary. Moreover, procedures to audit LLMs and other foundation models have yet to be developed. Therefore, it is useful to consider the merits and limitations of existing AI auditing procedures when applied to LLMs.

\subsection{Seven claims about auditing of LLMs}
As demonstrated above, a wide range of AI auditing procedures have already been developed.\footnote{For more comprehensive overviews of available AI auditing tools and procedures, see \cite{mokanderEthicsbased2021, metaxaAuditing2021,brownAlgorithm2021}.} However, not all auditing procedures are equally effective in handling the risks posed by LLMs. Nor are they equally likely to be implemented, due to factors including technical limitations, institutional access, and administrative costs \cite{mokanderEthics2021}. In what follows, we discuss some key distinctions that inform the design of auditing procedures and defend seven claims about making such designs feasible and effective for LLMs.

To start with, it is useful to distinguish between \textit{compliance audits} and \textit{risk audits}. The former compares an entity’s actions or properties to predefined standards or regulations. The latter asks open-ended questions about how a system works to identify and control risks. When conducting risk audits of LLMs, auditors can draw on well-established procedures, including standards for AI risk management \cite{nistAI2022, isoISO2023} and guidance on how to assess and evaluate AI systems \cite{nistRisk2002, vdeVDE2022, icoGuidance2020, iiaIIA2018, isoISO2022,floridiCapAI2022}. In contrast, compliance audits require a normative baseline against which AI systems can be evaluated. However, LLM research is a quickly developing field in which standards and regulations have yet to emerge. Moreover, the fact that LLMs are adaptable to many downstream applications undermines the feasibility of auditing procedures designed to ensure compliance with sector-specific norms and regulations. This leads to our first claim: 

\begin{framed}
\noindent \textbf{\hypertarget{claim1}{Claim 1.}} AI auditing procedures focusing on compliance alone are unlikely to provide adequate assurance for LLMs.
\end{framed}

Our blueprint for how to audit LLMs outlined in \cref{sec:4} accounts for (\hyperlink{claim1}{Claim 1}) by incorporating elements of both risk audits (at governance and model levels) and compliance audits (at the application level). 

Further, it is useful to distinguish between \textit{external} and \textit{internal audits}. The former is conducted by independent third parties and the latter by an internal function reporting directly to its board \cite{iiaInternal2022}. External audits help address concerns regarding accuracy in self-reporting \cite{sandvigAuditing2014}, so they typically underpin formal certification procedures \cite{yanisky-ravidEquality2019}. However, they are constrained by limited access to internal processes \cite{rajiClosing2020}. For internal audits, the inverse is true: while constituting an essential step towards informed model design decisions \cite{saleiroAequitas2019}, they run an increased risk of collusion between the auditor and the auditee \cite{costanza-chockWho2022}. Moreover, without third-party accountability, decision-makers may ignore audit recommendations that threaten their business interests \cite{sleeIncompatible2020}. The risks stemming from misaligned incentives are especially stark for technologies with rapidly increasing capabilities and for companies facing strong competitive pressures \cite{Naude2020}. Both conditions apply to LLMs, undermining the ability of internal auditing procedures to provide meaningful assurance in this space. This observation, combined with the need to manage the social and ethical risks posed by LLMs surveyed in \cref{sec:2}, leads us to assert that: 

\begin{framed}
\noindent \textbf{\hypertarget{claim2}{Claim 2.}} External audits are required to ensure that LLMs are ethical, legal, and technically robust. 
\end{framed}

As we explain in \cref{sec:4}, each step in our blueprint for how to audit LLMs should be conducted by independent third-party auditors. However, external audits come with their own challenges, including how to access information that is protected by privacy or IP rights \cite{englerOutside2021, rajiOutsider2022}. This is especially challenging in the case of LLMs since some are only accessible via an application programming interface (API) and others are not published at all. Determining the auditor’s level of access is thus an integral part of designing LLM auditing procedures.

Koshiyama et al. \cite{koshiyamaAlgorithm2022} proposed a typology that distinguishes between different access levels. At lower levels, auditors have no direct access to the model but base their evaluations on publicly available information about the development process. At middle levels, auditors have access to the computational model itself, meaning they can manipulate its parameters and review its task objectives. At higher levels, auditors have access equivalent to the system developer to all the details encompassing a system, i.e., full access to organisational processes, actual input and training data, and information about how and why the system was initially created. In \cref{sec:4}, we use this typology to indicate the level of access auditors need to conduct audits at the governance, model, and application levels.

The question about access leads us to a further distinction made in the AI auditing literature, i.e., between \textit{adversarial} and \textit{collaborative} audits. Adversarial audits are conducted by independent actors to assess the properties or impact an AI system has – without privileged access to its source code or technical design specifications \cite{metaxaAuditing2021}. Collaborative audits see technology providers and external auditors working together to assess and improve the process that shapes future AI systems’ design and safeguards \cite{berghoutAdvanced2023, mokanderEthics2021a}. While the former primarily aims to expose harms, the latter seeks to provide assurance. Previous research has shown that audits are most effective when technology providers and independent auditors collaborate towards the common goal of identifying and managing risks \cite{powerAudit1997}. This implies that: 

\begin{framed}
\noindent \textbf{\hypertarget{claim3}{Claim 3.}} To be feasible and effective in practice, procedures to audit LLM require active collaboration between technology providers and independent auditors.
\end{framed}

Accounting for (\hyperlink{claim3}{Claim 3}), this article focuses on collaborative audits. In fact, all steps in our three-layered approach outlined in \cref{sec:4} demand that technology providers provide external auditors with the access they need and proactively feed their own know-how into the process. After all, evaluating LLMs requires resources and technical expertise that technology providers are best positioned to provide. 

Moving on, it is also useful to distinguish between \textit{governance audits} and \textit{technology audits}. The former focus on the organisation designing or deploying AI systems and include assessments of software development and quality management processes, incentive structures, and the allocation of roles and responsibilities \cite{senftInformation2009}. The latter focus on assessing a technical system's properties, e.g., reviewing the model architecture, checking its consistency with predefined specifications, or repeatedly querying an algorithm and observing its outputs to understand its workings and potential impact \cite{metaxaAuditing2021}. Some LLM-related risks can be identified and mitigated at the application level. However, other issues are best addressed upstream, e.g., those concerning the sourcing of training data. This implies that, to be feasible and effective:

\begin{framed}
\noindent \textbf{\hypertarget{claim4}{Claim 4.}} Auditing procedures designed to assess and mitigate the risks posed by LLMs must include elements of both governance and technology audits.
\end{framed}

Our blueprint for how to audit LLMs satisfies this claim in the following way. The governance audits we propose aim to assess the processes whereby LLMs are designed and disseminated, the model audits focus on assessing the technical properties of pre-trained LLMs, and the application audits focus on assessing the technical properties of applications built on top of LLMs.

However, both governance and technology audits have limitations. During governance audits, for example, it is not possible to anticipate upfront all the risks that emerge as AI systems interact with complex environments over time \cite{lauerYou2020}. Further, not all ethical tensions stem from technology design alone, as some are intrinsic to specific tasks or applications \cite{danksRegulating2017}. While these limitations of governance audits are well-known, highly general AI systems introduce new challenges for technology audits, which have historically focused on assessing systems designed to fill specific functions in well-defined contexts, e.g., improving image analysis in radiology \cite{mahajanAlgorithmic2020} or detecting corporate fraud \cite{zerbinoProcessminingenabled2018} However, because LLMs enable many downstream applications, traditional auditing procedures cannot capture the full range of LLM-related risks. While existing best practices in governance auditing appear applicable to organisations designing or deploying LLMs, that is not true for technology audits. In short:

\begin{framed}
\noindent \textbf{\hypertarget{claim5}{Claim 5.}} The methodological design of technology audits will require significant modifications to identify and assess LLM-related risks.
\end{framed}

As mentioned above, our blueprint for how to audit LLMs incorporates elements of technology audits on both the model and the application levels. To understand why that is necessary to identify and mitigate the ethical risks posed by LLMs, we must first distinguish between different types of technology audits. 

Previous work on technology audits distinguish between \textit{functionality}, \textit{model}, and \textit{impact} audits \cite{mittelstadtAuditing2016}. Functionality audits focus on the rationale underpinning AI systems by asking questions about intentionality, e.g., what is this system's purpose \cite{krollFallacy2018}? Model audits review the system's decision-making logic. For symbolic AI systems,\footnote{Symbolic AI systems are based on explicit methods like first-order logic and decision trees. Sub-symbolic systems rely on establishing correlations through statistical methods like Bayesian learning and back-propagation \cite{russellArtificial2015}.} that entails reviewing the source code. For sub-symbolic AI systems, including LLMs, it entails asking how the model was designed, the training data used, and performance on different benchmarks. Finally, impact audits investigate the types, severity, and prevalence of effects from an AI system's outputs on individuals, groups, and the environment \cite{oecdOECD2022}. These approaches are not mutually exclusive but rather highly complementary \cite{mokanderEthics2021a}. Still, the fact that LLM developers have limited information about the future deployment of their systems, leads us to make the following claim:

\begin{framed}
\noindent \textbf{\hypertarget{claim6}{Claim 6.}} Model audits will play a key role in identifying and communicating LLMs' limitations, thereby informing system redesign, and mitigating downstream harm.
\end{framed}

This claim constitutes a key justification for the three-layered approach to LLM auditing proposed in this article. As highlighted in \cref{sec:4}, governance audits and application audits are both well-established practices in systems engineering and software development. Hence, it is precisely by adding structured and independent audits on the model level that our blueprint for auditing LLMs complements and enhances existing governance structures.

Finally, within technology audits, it is important to distinguish between \textit{ex-ante} and \textit{ex-post} audits, which take place before and after a system is deployed, respectively. The former can identify and prevent some harms before they occur while informing downstream users about the model's appropriate, intended applications. Considerable literature already exists within computer science on techniques such as model fooling \cite{xuFooling2018}, functional testing \cite{rottgerHateCheck2021} and template-based stress-testing \cite{aspillagaStress2020}, which all play important roles during technology audits of LLMs. However, ex-ante audits cannot fully capture all the risks associated with systems that continue to ‘learn' by updating their internal decision-making logic \cite{dignumResponsible2017}.\footnote{In their unfrozen states, all LLMs can learn as they are fed new data. However, once a model has been `fixed', it does not update and simply uses new input data to make predictions.} This limitation applies to all learning systems but is particularly relevant for LLMs that display emergent properties \cite{weiEmergent2022a}.\footnote{Emergence implies that an entity can have properties its parts do not individually possess, and that randomness can generate orderly structures \cite{corningReemergence2010}.} Ex-post audits can be divided into \textit{snapshot audits} (which occur once or on regular occasions) and \textit{continuous audits} (which monitor performance over time). Most existing AI auditing procedures are snapshots.\footnote{The post-market monitoring plans mandated in the proposed EU AI Act \cite{commissionArtificial2021} is a rare example of continuous auditing.}  Like ex-ante audits, however, snapshots are unable to provide meaningful assurance regarding LLMs as they display emergent capabilities and, in some cases, can learn as they are fed new data. This leads to our final claim: 

\begin{framed}
\noindent \textbf{\hypertarget{claim7}{Claim 7.}} To be effective, LLM auditing procedures must include some elements of continuous ex-post auditing.
\end{framed}

In our blueprint, continuous ex-post monitoring is one of the activities conducted at the application level. However, as detailed in \cref{sec:4}, audits on the different levels are strongly interconnected. For example, continuous monitoring of LLM-based applications presupposes that technology providers have established ex-post monitoring plans – which can only be verified by audits at the governance level. Invertedly, technology providers rely on feedback from audits at the application level to continue improving their software development and quality management procedures.

To summarise, much can be learned from existing AI auditing procedures. However, LLMs display several properties that undermine the feasibility of such procedures. Specifically, LLMs are adaptable to a wide range of downstream applications, display emergent capabilities, and can, in some cases, continue to learn over time. As this section has shown, that means that neither functionality audits (which hinge on the evaluation of the purpose of a specific application) nor impact audits (which hinge on the ability to observe a specific system’s actual impact) alone can provide meaningful assurance against the social and ethical risks LLMs pose. It also means that ex-ante audits must be complemented by continuous post-market monitoring of outputs from LLM-based applications.

In this section, we have built on these and other insights to derive and defend seven claims about how auditing procedures should be designed to account for the governance challenges LLMs pose. These seven claims provided our starting point when designing the three-layered approach for auditing LLMs that will be outlined in \cref{sec:4}. However, we maintain that these claims are more general and could serve as guardrails for other attempts to design auditing procedures for all foundation models.

%% file: sections/4_auditing_LLMs.tex
\section{Auditing LLMs: A three-layered approach}
\label{sec:4}
This section offers a blueprint for auditing LLMs that satisfies the seven claims in Sec 3 about how to structure such procedures. While there are many ways to do that, our proposal focuses on a limited set of activities that are (i) jointly sufficient to identify LLM-related risks, (ii) practically feasible to implement, and (iii) have a justifiable cost-benefit ratio. The result is the three-layered approach outlined below.

\subsection{A blueprint for LLM auditing}

\begin{figure*}[!b]
    \centering
    \includegraphics[width = \textwidth]{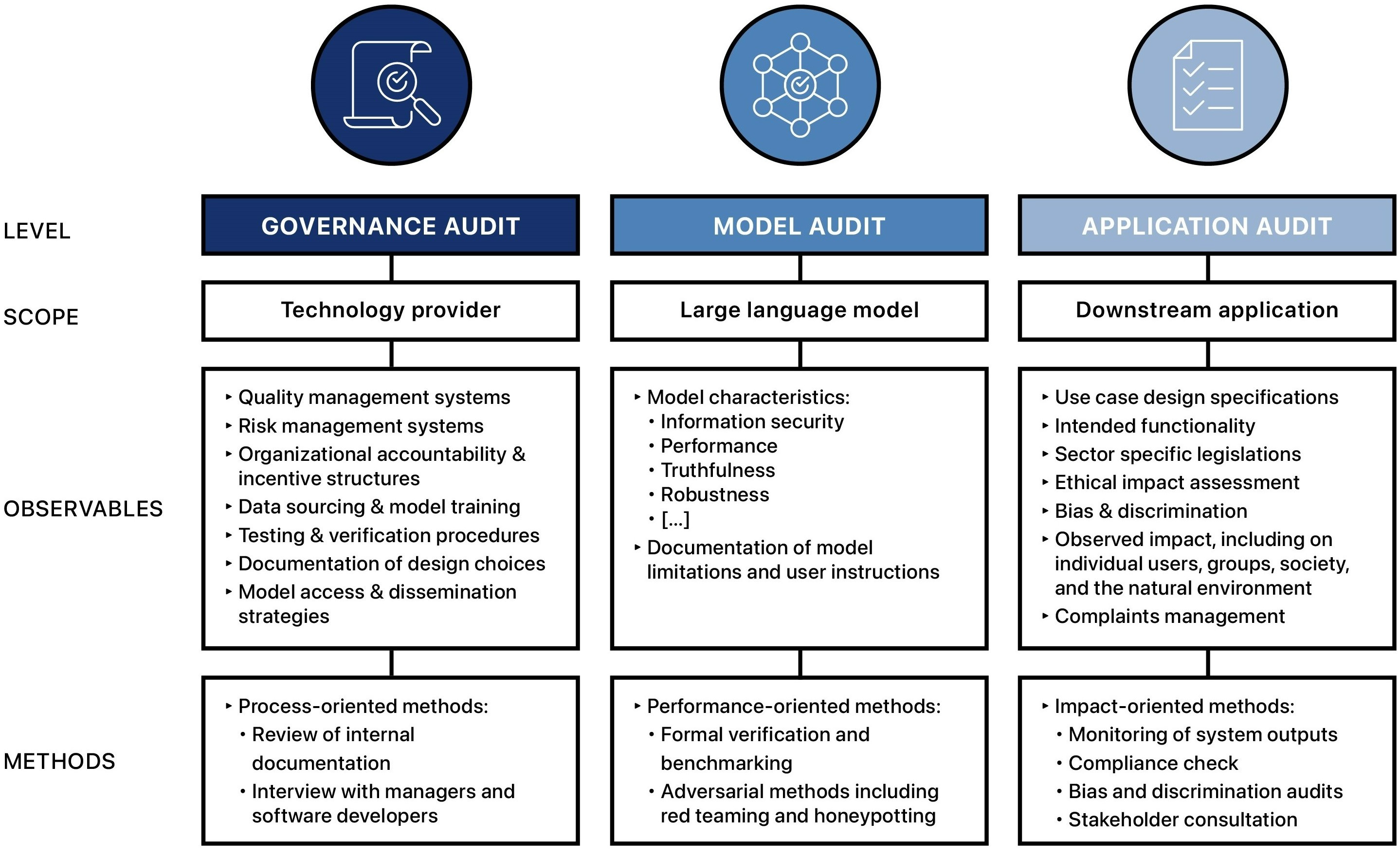}
    \caption{Blueprint for how to audit LLMs: A three-layered approach.}
    \label{fig:1}
\end{figure*}

Audits should focus on three levels. First, technology providers developing LLMs should undergo \textit{governance audits} that assess their organisational procedures, accountability structures and quality management systems. Second, LLMs should undergo \textit{model audits}, assessing their capabilities and limitations after initial training but before adaptation and deployment in specific applications. Third, downstream applications using LLMs should undergo continuous \textit{application audits} that assess the ethical alignment and legal compliance of their intended functions and their impact over time. \cref{fig:1} illustrates the logic of our approach.

Some clarifications are needed to flesh out our blueprint. To begin with, governance, model and application audits only provide effective assurance when coordinated. As \cref{sec:3} showed, LLM audits must include elements of process- and performance-oriented auditing (\hyperlink{claim4}{Claim 4}). In our three-layered approach, the governance audits are process-oriented, whereas the model and application audits are performance-oriented. As previously discussed, feasible and effective LLM auditing procedures must include aspects of continuous, ex-post assessments (\hyperlink{claim7}{Claim 7}). In our blueprint, these elements are incorporated at the application level. But this is just two examples. As we discuss what governance, model and applications audits entail in this section, we also make highlight how they, when combined, satisfy all seven claims listed in \cref{sec:3}.

While audit types in our blueprint are individually necessary, their boundaries overlap and can be drawn in multiple ways. For example, the collection and pre-processing of training data ties into software development practices. Hence, reviewing organisational procedures for obtaining and curating training data is legitimate during holistic governance audits. However, LLMs inherently reflect biases in their training data \cite{shengWoman2019, Gehman2020}.\footnote{The link between model characteristics and biases in the training data can be counterintuitive \cite{molnar2018interprtable}. In some cases, biased datasets can help models recognise bias and steer away from it with the help of reinforcement learning and human feedback \cite{Christiano2017}.} Reviewing such data is, therefore, often necessary in model audits. Nevertheless, the conceptual distinction between governance, model and application audits remains useful when identifying varied risks that LLMs pose.

It is theoretically possible to add further layers to our blueprint. For example, downstream developers could also be made subject to process-oriented governance audits. But such audits would be difficult to implement, given that many decentralised actors build applications on top of LLMs. The combination of governance, model, and application audits, we argue, strikes a balance between covering a sufficiently large part of the development and deployment lifecycle to identify LLM-related risks, on the one hand, and being practically feasible to implement, on the other. Regardless of how many layers are included, however, the success of our blueprint relies on responsible actors at each level who actively want to or are incentivised to ensure good governance.

Finally, to provide meaningful assurance, audits on all three levels should be external (\hyperlink{claim2}{Claim 2}) yet collaborative (\hyperlink{claim3}{Claim 3}). In practice, this implies that independent third parties not only seek to verify claims made by technology providers but also work together with them to identify and mitigate risks and shape the design of future LLMs. As mentioned in the introduction, the question of who should conduct the audits falls outside the scope of this article. That said, reasonable concerns about how independent collaborative audits really are can be raised regardless of who is conducting the audit. In \cref{sec:5}, we discuss this and other limitations.

With those clarifications in mind, we will now present the details of our three-layered approach. The following subsections discuss governance, model, and application audits, focusing on why each is needed, what each entails, and what outputs each should produce. 

\subsection{Governance audits}
Technology providers working on LLMs should undergo governance audits that assess their organisational procedures, incentive structures, and management systems. Overwhelming evidence shows that such features influence the design and deployment of technologies \cite{brundageTrustworthy2020a}. Moreover, research has demonstrated that risk-mitigation strategies work best when adopted transparently, consistently, and with executive-level support \cite{floridiEthical2020, hodgesEthics2015}. Technology providers are responsible for identifying the risks associated with their LLMs and are uniquely well-positioned to manage some of those risks. Therefore, it is crucial that their organisational procedures and governance structures are adequate.

Governance audits have a long history in areas like IT governance \cite{senftInformation2009, isoISO2015,iliescuAuditing2010} and systems and safety engineering \cite{falcoGoverning2021, levesonEngineering2011, dobbeSystem2022}. Tasks include assessing internal governance structures, product development processes and quality management systems \cite{berghoutAdvanced2023} to promote transparency and procedural regularity, ensure that appropriate risk management systems are in place \cite{schuettThree2022a}, and spark deliberation regarding ethical and social implications throughout the software development lifecycle. Governance audits can also improve accountability, e.g., publicising their results prevents companies from covering up undesirable outcomes and incentivises better behaviour \cite{englerOutside2021}. Thus defined, governance audits incorporate elements of both compliance audits, regarding completeness and transparency of documentation, and risk audits, regarding the adequacy of the risk management system (\hyperlink{claim1}{Claim 1}).

We argue that governance audits of LLM providers should focus on three tasks:\footnote{Governance audits could examine many tasks, and prioritization may vary depending on the sector and jurisdiction. Hence, the three tasks we propose are merely a minimum baseline.}

\begin{enumerate}
    \item \textit{Reviewing the adequacy of organisational governance structures} to ensure that model development processes follow best practices and that quality management systems can capture LLM-specific risks. While technology providers have in-house quality management experts, confirmation bias may prevent them from recognising critical flaws; involving external auditors addresses that issue \cite{bauerNecessity2016}. Nevertheless, governance audits are most effective when auditors and technology providers collaborate to identify risks \cite{chopraSociotechnical2018}. Therefore, it is important to distinguish accountability from blame at this stage of an audit.
    \item \textit{Creating an audit trail of the LLM development process} to provide chronological documentary evidence of the development of an LLM's capabilities, including information about its intended purpose, design specifications and choices, as well as training data and testing procedures through the generation of model cards \cite{mitchellModel2019}, datasheets \cite{gebruDatasheets2021} and system cards \cite{metaaiSystem2023}.\footnote{Meta AI's detailed notes on the OPT model provide an exemplar of model training documentation \cite{zangMetaseq2022}.} Audit trails serve several related purposes. Stipulating design specifications upfront facilitates checking system adherence to jurisdictional requirements downstream \cite{falcoGoverning2021}. Similarly, information concerning intended use cases should inform licensing agreements with downstream developers \cite{contractorBehavioral2022}, thereby restricting the potential for harm through misuse. Finally, requiring providers to document and justify their design choices sparks ethical deliberation by making trade-offs explicit.
    \item \textit{Mapping roles and responsibilities within organisations that design LLMs} facilitates the allocation of accountability for system failures. LLMs' adaptability downstream does not exculpate technology providers from all responsibility. Some risks are `reasonably foreseeable'. In the adjacent field of machine learning (ML) image recognition, a study found that commercial gender classification systems were less accurate for darker-skinned females than lighter-skin males \cite{buolamwiniGender2018}. After the release of these findings, all technology providers speedily improved the accuracy of their models, suggesting that the main problem was intrinsic, but resulted from inadequate risk management. Mapping the roles and responsibilities of different stakeholders improves accountability and increases the likelihood of impact assessments being structured rather than ad-hoc, thus helping identify and mitigate harms proactively. 
\end{enumerate}

To conduct these three tasks, auditors primarily require what Koshiyama et al. \cite{koshiyamaAlgorithm2022} refer to as white-box auditing. This is the highest level of access and suggests that the auditor knows how and why an LLM was developed. In practice, it implies privileged access to facilities, documentation, and personnel, which is standard practice in governance audits in other fields. For example, IT auditors have full access to material and reports related to operational processes and performance metrics \cite{senftInformation2009}. It also implies access to the input data, learning procedures, and task objectives used to train LLMs. White-box auditing requires that nondisclosure and data-sharing agreements are in place, which adds to the logistical burden of governance audits. However, granting such a high level of access is especially important from an AI safety perspective because, in addition to auditing LLMs before market deployment, governance audits should also evaluate organisational safeguards concerning high-risk projects that providers may prefer not to discuss publicly.

The results of governance audits should be provided in formats tailored to different audiences. The primary audience is the management and directors of the LLM provider. Auditors should provide a full report that directly and transparently lists and discusses the vulnerabilities of existing governance structures. Such reports may recommend actions, but taking actions remains the provider’s responsibility. Usually, such audit reports are not made public. However, some evidence obtained during governance audits can be curated for two secondary audiences: law enforcers and developers of downstream applications. In some jurisdictions, hard legislation may demand that technology providers follow specific requirements. For instance, the proposed EU AI Act required providers to register high-risk AI systems with a centralised database \cite{commissionArtificial2021} or implement a risk management system \cite{schuettRisk2023}. In such cases, reports from independent governance audits can help providers demonstrate adherence to legislation. Reports from governance audits also assist developers of downstream applications to understand an LLM’s intended purpose, testing, verification, risks, and limitations.

Before concluding this discussion, it is useful to reflect on how governance audits contribute to relieving some of the social and ethical risks LLMs pose. As mentioned in Sec 2, Weidinger et al. \cite{weidingerEthical2021b} listed six broad risk areas: discrimination, information hazards, misinformation hazards, malicious use, human-computer interaction harm, and automation and environmental harms. Governance audits address some of these directly. By assessing the adequacy of the governance structures surrounding LLMs, including licencing agreements \cite{contractorBehavioral2022} and structured access protocols \cite{shevlaneStructured2022}, governance audits help reduce the risk of malicious use. Further, some information hazards stem from the possibility of extracting sensitive information from LLMs via adversarial attacks \cite{carliniExtracting2021}. By reviewing the process whereby training datasets were sourced, labelled, and curated, as well as the strategies and techniques used during the model training process – such as differential privacy \cite{Dwork2006} or secure federated learning \cite{kaissis2020secure} – governance audits can minimise the risk of LLMs leaking sensitive information. However, for most of the risk areas listed by Weidinger et al. \cite{weidingerEthical2021b}, governance audits have only an indirect impact insofar as they contribute to transparency about the limitations and intended purposes of LLMs. Hence, risks areas like discrimination, misinformation hazards, and human-computer interaction harms are better addressed by model and application audits. 

\subsection{Model audits}
Before deployment, LLMs should be subject to model audits that assess their capabilities and limitations (\hyperlink{claim6}{Claim 6}). Model audits share some features with governance audits. For example, both happen before an LLM is adapted for specific applications. However, model audits do not focus on organisational procedures but on LLMs' capabilities and characteristics. Specifically, they should identify an LLM's limitations to (i) inform the continuous redesign the system, and (ii) communicate its capabilities and limitations to external stakeholders. These two tasks use similar methodologies, but they target different audiences.

The first task -- limitation identification -- aims primarily to support organisations that develop LLMs with benchmarks or other data points that inform internal model redesigning and retraining efforts \cite{bharadwajPattern2021}. Model audits' results should also inform API license agreements, helping prevent applications in unintended use cases \cite{contractorBehavioral2022} and restricting the distribution of dangerous capabilities \cite{shevlaneStructured2022}. The second task -- communicating capabilities and limitations -- aims to inform the design of specific applications built on top of LLMs by downstream developers. Such communication can take different forms, e.g., \textit{interactive model cards} \cite{crisanInteractive2022} and \textit{information about the initial training dataset} \cite{pushkarnaData2022,jerniteData2022}, to help downstream developers adapt the model appropriately.

In \cref{sec:3}, we argued that the way technology audits are being conducted requires modifications to address the governance challenges associated with LLMs (\hyperlink{claim5}{Claim 5}). In what follows, we demonstrate that evaluating an LLM’s characteristics independent of an intended use case is challenging but not impossible.\footnote{A wide range of tools and methods to evaluate LLMs already exists. For an overview, see the report \textit{Holistic Evaluation of Language Models} published by researchers at the Center for Research on Foundation Models \cite{liangHolistic2022a}.} To do so, auditors can use two distinct approaches. The first involves identifying and assessing intrinsic characteristics. For example, the training dataset can be assessed for completeness and consistency without reference to specific use cases \cite{floridiCapAI2022}. However, it is often expensive and technically challenging to interrogate large datasets \cite{Paullada2021}. The second involves employing an indirect approach that tests the model across multiple potential downstream use cases, links the results to different characteristics, and assesses the aggregated results using different weighting techniques. That second approach may prove more fruitful when assessing an LLM’s performance.

Nevertheless, selecting the characteristics to focus on during model audits remains challenging. Given such audits' purpose, we recommend examining characteristics that are:
\begin{itemize}
    \item \textit{Socially and ethically relevant}, i.e., that can be directly linked to the social and ethical risks posed by LLMs;
    \item \textit{Predictably transferable}, i.e., that impact the nature of downstream applications; and
    \item \textit{Meaningfully operationalisable}, i.e., that can be assessed with the available tools and methods. 
\end{itemize}

Keeping those criteria in mind, we posit that model audits should focus on (at least) the performance, robustness, information security and truthfulness of LLMs. As other characteristics may meet the three criteria listed above, those four characteristics are just examples highlighting the role of model audits in our three-layered approach. The list of relevant model characteristics can be amended as required when developing specific auditing procedures.

With those caveats out of the way, we now proceed to discuss how four example characteristics can be assessed during model audits:

\begin{enumerate}
    \item \textit{Performance, i.e., how well the LLM functions on various tasks.} Standardised benchmarks can help assess an LLM's performance by comparing it to a human baseline. For example, \textit{GLUE} \cite{wangGLUE2018} aggregates LLM performance across multiple tasks into a single reportable metric. Such benchmarks have been criticised for overestimating performance over a narrow set of capabilities and quickly becoming saturated, i.e., rapidly converging on the performance of non-expert humans, leaving limited space for valuable comparisons. Therefore, it is crucial to evaluate LLMs' performance against many tasks or benchmarks, and sophisticated tools and methods have been proposed for that purpose, including \textit{SuperGLUE} \cite{wangSuperGLUE2019}, which is more challenging and `harder to game' with narrow LLM capabilities, and \textit{BIG-bench} \cite{srivastavaImitation2022a}, which can assess LLM's performance on tasks that appear beyond their current capabilities. These benchmarks are particularly relevant for model audits because they were primarily developed to evaluate pre-trained models, without task-specific fine-tuning.
    \item \textit{Robustness, i.e., how well the model reacts to unexpected prompts or edge cases.} In ML, robustness indicates how well an algorithm performs when faced with new, potentially unexpected (i.e., out-of-domain) input data. LLMs lacking robustness introduce, at least, two distinct risks \cite{rudnerKey2021}. First, the risk of critical system failures if, for example, an LLM performs poorly for individuals, unlike those represented in the training data \cite{Sohoni2020}. Second, the risk of adversarial attacks \cite{gargBAE2020,liBERTATTACK2020}. Therefore, researchers and developers have created tools and methods to assess LLMs' robustness, including evaluation toolkits like the \textit{Robustness Gym} \cite{goelRobustness2021}, benchmark datasets like \textit{ANLI} \cite{nieAdversarial2020}, and open-source platforms like \textit{Dynabench} \cite{kielaDynabench2021}. Particularly relevant for our purposes is \textit{AdvGLUE} \cite{wangAdversarial2021}, which evaluates LLMs' vulnerabilities to adversarial attacks in different domains using a multi-task benchmark. By quantifying robustness, AdvGLUE facilitates comparisons between LLMs and their various affordances and limitations. However, robustness can be operationalised in different ways, e.g., group robustness, which measures a model's performance across different sub-populations \cite{zhangContrastive2022}. Therefore, model audits should employ multiple tools and methods to assess robustness.
    \item \textit{Information security, i.e., how difficult it is to extract training data from the LLM.} Several LLM-related risks can be understood as `information hazards' \cite{weidingerEthical2021b}, including the risk of compromising privacy by leaking personal data. As demonstrated by \cite{carliniExtracting2021}, adversarial agents can perform \textit{training data extraction attacks} to recover personal information like names and social security numbers. However, not all LLMs are equally vulnerable to such attacks. The memorisation of training data can be minimised through differentially private training techniques \cite{mcmahanLearning2018}, but their application generally reduces accuracy \cite{Jayaraman2019} and increases training time \cite{songAuditing2019}. Promisingly, it is possible to assess the extent to which an LLM has unintentionally memorised rare or unique training data sequences using metrics such as \textit{exposure} \cite{Carlini2019}. Testing strategies, like exposure, can be employed at the model level, although that requires auditors to have access to the LLM and its training corpus. Still, assessing LLMs' information security during model audits does not address all information hazards because some risk of correctly inferring sensitive information about users can only be audited on an application level.
    \item \textit{Truthfulness, i.e., to what extent the LLM can distinguish between the real world and possible worlds.} Some LLM-related risks stem from their capacity to provide false or misleading information, which creates less well-informed users and potentially erodes public trust in shared information \cite{weidingerEthical2021b}. Statistical methods struggle to distinguish between factually correct versus plausible but factually incorrect information. That problem is exacerbated by the fact that many LLM training practices, like imitating human text on the web or optimising for clicks, are unlikely to create truthful AI \cite{evansTruthful2021}.\footnote{Alternative techniques that are better suited for developing truthful AI include bootstrapping, adversarial training \cite{hubingerRelaxed2019} and transparent AI \cite{wellerChallenges2017}.} However, during model audits, our concern is not developing truthful AI but evaluating truthfulness. Such audits should focus on evaluating overall truthfulness, not the truthfulness of an individual statement. However, that does not preclude focusing on multiple aspects, e.g., how frequent falsehoods are on average, and how bad worst-case falsehoods are. One benchmark that measures truthfulness is \textit{TruthfulQA} \cite{linTruthfulQA2022}, which generates a percentage score using 817 questions spanning 38 application domains, including healthcare and politics. When evaluating an LLM with the help of TruthfulQA, auditors would get a percentage score on how truthful the model is. However, even a strong performance on TruthfulQA does not imply that an LLM will be truthful in a specialised domain. Nevertheless, such benchmarks offer helpful tools for model audits. 
\end{enumerate}

These four characteristics pertain to pre-trained LLMs. However, model audits should also review training datasets. It is well-known that training data gaps or biases create models that perform poorly on different datasets \cite{nejadgholiCrossdataset2020}. Training LLMs with biased or incomplete data can cause representational and allocational harms \cite{caliskanSemantics2017}. Therefore, a recent European Parliament report \cite{eprsAuditing2022} discussed mandating third-party audits of AI-training datasets. Technology providers should prepare for such suggestions potentially becoming legal requirements.

Despite these technical and legal considerations, training datasets are often collected with little curation, supervision, or foresight \cite{joLessons2020}. While curating `unbiased' datasets may be impossible, disclosing how a dataset was assembled can suggest its potential biases \cite{dodgeDocumenting2021}. Model auditors can use existing tools and methods that interrogate biases in LLMs' pre-trained word embeddings, such as the metrics \textit{DisCo} \cite{websterMeasuring2021}, \textit{SEAT} \cite{mayMeasuring2019} or \textit{CAT} \cite{nadeemStereoSet2021}. So-called \textit{data statements} \cite{benderData2018} can provide developers and users with the context required to understand specific models' potential biases. \textit{Data representativeness criterion} \cite{schatData2020} can determine how representative\footnote{The term `representativeness' has different meaning in statistics, politics, and machine learning, ranging from a proportionate match between sample and population to a more general sense of inclusiveness \cite{chasalowRepresentativeness2021}.} a training dataset is, and manual datasets audits can be supplemented with \textit{automatic analysis} \cite{kreutzerQuality2022}. The \textit{Text Characterisation Toolkit} \cite{simigText2022} permits automatic analysis of how dataset properties impact model behaviour. While the availability of such tools is encouraging, it is important to remain realistic about what dataset audits can achieve. To reiterate, model audits do not aim to ensure that LLMs are ethical in any global sense. Instead, they contribute to better precision in claims about an LLM's capabilities and inform the design of downstream applications.

Model audits require auditors to have privileged access to LLMs and their training datasets. In the typology provided by Koshiyama et al. \cite{koshiyamaAlgorithm2022}, this corresponds to medium-level access, whereby auditors have access to an LLM equivalent to its developer, meaning they can manipulate model parameters and review learning procedures and task objectives. Such access is required to assess LLMs' capabilities accurately during model audits. However, in contrast to white-box audits, the access model auditors enjoy is limited to the technical system and does not extend to technology providers’ organisational processes.

Some of the characteristics tested for during model audits correspond directly to the social and ethical risks LLMs pose. For example, our model audits entail evaluating LLMs according to characteristics like information security and truthfulness, which correspond to information hazards and misinformation hazards, respectively, in Weidinger et al.’s taxonomy \cite{weidingerEthical2021b}. However, it should be noted that our proposed model audits only focus on a few characteristics of LLMs. That is because the criterion of meaningful operationalisability sets a high bar: not all risks associated with LLMs can be addressed at the model level. Consider discrimination as an example. Model audits can expose the root causes of some discriminatory practices, such as biases in training datasets that reflect historic injustices. However, what constitutes unjust discrimination is context-dependent and varies between jurisdictions. That problematises saying anything meaningful about risks like unjust discrimination on a model level \cite{hancox-liEpistemic2021}. However, that observation does not argue against model audits but for complementary approaches like application audits, as discussed next.

\subsection{Application audits}
Products and services built using LLMs should undergo application audits that assess the legality of their intended functions and how they will impact users and societies. Unlike governance and model audits, application audits focus on actors employing LLMs in downstream applications. Such audits are well-suited to ensure compliance with national and regional legislation, sector-specific standards, and organisational ethics principles. Application audits have two components: \textit{functionality audits}, which evaluate applications using LLMs based on their intended and operational goals, and \textit{impact audits}, which evaluate applications based on their impacts on different users, groups, and the natural environment. Both are well-established in AI auditing. They are also crucially complementary \cite{dashNetworkcentric2019}. Next, we consider how they can be combined into procedures for auditing applications based on LLMs.

During \textit{functionality audits}, auditors should check whether the intended purpose of a specific application is (1) legal and ethical in and of itself and (2) aligned with the intended use of the LLM in question. The first check is for legal and ethical compliance, i.e., the adherence to the laws, regulations, guidelines, and specifications relevant to a specific application \cite{idowuLegal2013}, as well as to voluntary ethics principles \cite{jobinGlobal2019} or codes of conduct \cite{greenMethods2009}. The purpose of these compliance checks is straightforward: if an application is unlawful or unethical, the performance of its LLM component is irrelevant, and the application should not be permitted on the market.

The second check within functionality audits also aim to address the risks stemming from developers overstating or misrepresenting a specific application's capabilities \cite{rajiFallacy2022}. During governance audits, technology providers are obliged to define the intended and disallowed use cases of their LLMs. During model audits, the limitations of LLMs are documented to inform their adaptation downstream. Using such information, functionality audits should ensure that downstream applications are aligned with a given LLM’s intended use cases in ways that take account of the model’s limitations. Functionality audits thus combines the elements of compliance and risks audit needed to provide assurance for LLMs (\hyperlink{claim1}{Claim 1}).

During \textit{impact audits}, auditors disregard an application's intended purpose and technological design to focus only on how its outputs impact different user groups and the environment. The idea behind impact audits is simple: every system can be understood in terms of its inputs and outputs \cite{krollFallacy2018}. However, despite that simplicity, implementing impact audits is notoriously hard. To begin with, AI systems and their environments co-evolve in non-linear ways \cite{lauerYou2020}. Therefore, the link between an LLM-based application's intended purpose and its actual impact may be neither intuitive nor consistent over time. Moreover, it is difficult to track impacts stemming from indirect causal chains \cite{rahwanSocietyintheloop2018, dafoeAI2017}. Consequently, establishing which direct and indirect impacts are considered legally and socially relevant remains a context-dependent question which must be resolved on a case-by-case basis. The application must be redesigned or terminated if the impact is considered unacceptable.

Importantly, impact audits should include both pre-deployment (ex-ante) assessments and post-deployment (ex-post) monitoring (\hyperlink{claim7}{Claim 7}).\footnote{This structure mirrors the `conformity assessments' and ‘post-market monitoring plans' proposed in the EU AI Act \cite{mokanderConformity2022}.} The former can leverage either empirical evidence or plausible scenarios, depending on how well-defined the application is and the predictability of the environments in which it will operate. For example, applications can be tested in \textit{sandbox environments} \cite{trubySandbox2022} that mimic real-world environments and allow developers and policymakers to understand the potential impact before an application goes to market. When used for ML-based systems, sandboxes have proven safe harbours in which to detect and mitigate biases \cite{akpinarSandbox2022a}. However, real-world environments often differ from training and testing environments in unforeseen ways \cite{zindaEthics2021}. Hence, pre-deployment assessments of LLM-based applications must also use analytical strategies to anticipate the application's impact, e.g., \textit{ethical impact assessments} \cite{manteleroAI2018,reismanAlgorithmic2018, selbstInstitutional2021} and \textit{ethical foresight analysis} \cite{floridiEthical2020}.

Pre-deployment impact assessments and post-deployment monitoring are individually necessary, but they are not jointly sufficient. As policymakers are well-aware, capturing the full range of potential harms from LLM-based applications requires auditing procedures to include elements of continuous oversight (again, see \hyperlink{claim7}{Claim 7}). For example, the EU AI Act requires technology providers to document and analyse high-risk AI systems' performance throughout their life cycles \cite{commissionArtificial2021}. Methodologically, post-deployment monitoring can be done in different ways, e.g., periodically reviewing the output from an application and comparing it to relevant standards. Such procedures can also be automated, e.g., by using oversight programs \cite{etzioniAI2016} that continuously monitor and evaluate system outputs and alert or intervene if they transgress predefined tolerance spans. Such monitoring can be done by both private companies and government agencies \cite{whittlestoneAI2022}. Overall, application audits seek to ensure that ex-ante testing and impact assessments have been conducted following existing best practices; that post-market plans have been established to enable continuous monitoring of system outputs; and that procedures are in place to mitigate or report different types of failure modes. 

By focusing on individual use cases, application audits are well-suited to alerting stakeholders to risks that require much contextual information to understand and address. This includes risks related to discrimination and human-computer interaction harms in Weidinger et al.’s taxonomy \cite{weidingerEthical2021b}. Application audits help identify and manage such risks in several different ways. For example, quantitative assessments linking prompts with outputs can give a sense of what kinds of language an LLM is propagating and how appropriate that communication style and content is in different settings \cite{Karan2019, Gao2017}. Moreover, qualitative assessments (e.g., those based on interviews and ethnographic methods) can provide insights into users’ lived experiences of interacting with an LLM \cite{mardaImportance2021}. 

However, despite those methodological affordances, it remains difficult to define some forms of harm in any global sense \cite{delobelleMeasuring2022}. For example, several studies have documented situations in which LLMs propagate toxic language \cite{Gehman2020,nozzaHONEST2021}, but the interpretation of toxicity and the materialisation of its harms vary across cultural, social, or political groups \cite{costelloSocial2019, sapAnnotators2022, kirkHandling2022a}. Sometimes, `detoxifying' an LLM may be incompatible with other goals and potentially suppress texts written about or by marginalised groups \cite{welblChallenges2021}. Moreover, certain expressions might be acceptable in one setting but not in another. In such circumstances, the most promising way forward is to audit not LLMs themselves but downstream applications -- thereby ensuring that each application's outputs adhere to contextually appropriate conversational conventions \cite{kasirzadehConversation2022c}.

Another example concerns harmfulness, i.e., the extent to which an LLM-based application inflicts representational, allocational or experiential harms.\footnote{\cite{blodgettLanguage2020} distinguish between representational harms (portraying some groups more favourably than others) and allocation harms (allocating resources or opportunities unfairly by social group).} An LLM that lacks robustness or performs poorly for some social groups may permit unjust discrimination \cite{weidingerEthical2021b} or violate capability fairness \cite{rauhCharacteristics2022a} when informing real-world allocational decisions like hiring. Multiple benchmarks exist to assess model stereotyping of social groups, including \textit{CrowS-Pairs} \cite{nangiaCrowSPairs2020}, \textit{StereoSet} \cite{nadeemStereoSet2021} or \textit{Winogender} \cite{rudingerGender2018a}. To assess risks from experiential harms, quantitative assessments of LLM outputs give a sense of the language it is propagating. For example, \cite{Gehman2020} have developed the \textit{RealToxicityPrompts} benchmark to assess the toxicity of a generated completion.\footnote{This benchmark relies on \textit{PerspectiveAPI} to score `toxicity', which is a limitation given that system's weaknesses \cite{kirkHatemoji2022,kumarDesigning2021}.} However, the tools mentioned above are only examples. The main point here is that representational, allocational and experiential harms associated with LLMs are best assessed at the application level through functionality and impact audits as described in this section.

To conduct application audits, lower levels of access are typically sufficient. For example, to make quantitative assessments to determine the relationship between inputs and outputs, it is sufficient that auditors have what Koshiyama et al. \cite{koshiyamaAlgorithm2022} refer to as black box model access or, in some cases, input data access. Similarly, to audit LLM-based applications for legal compliance and ethical alignment, auditors do not require direct access to the underlying model but can rely on publicly available information – including the claims technology providers and downstream developers make about their systems and the user instructions attached to them.

We contend that governance audits and model audits should be obligatory for all technology providers designing and disseminating LLMs. However, we recommend that application audits should be employed more selectively. Further, although application audits may form the basis for certification \cite{cihonAI2021}, auditing does not equal certification. Certification requires predefined standards against which a product or service can be audited and institutional arrangements to ensure the certification process's integrity \cite{yanisky-ravidEquality2019}. Even when not related to certification, application audits' results should be publicly available (at least in summary form). Registries publishing such results incentivise companies to correct behaviour, inform enforcement actions and help cure informational asymmetries in technology regulation \cite{rajiOutsider2022}.

\subsection{Connecting the dots}
\begin{figure}
    \centering
    \includegraphics[width = 0.9\textwidth]{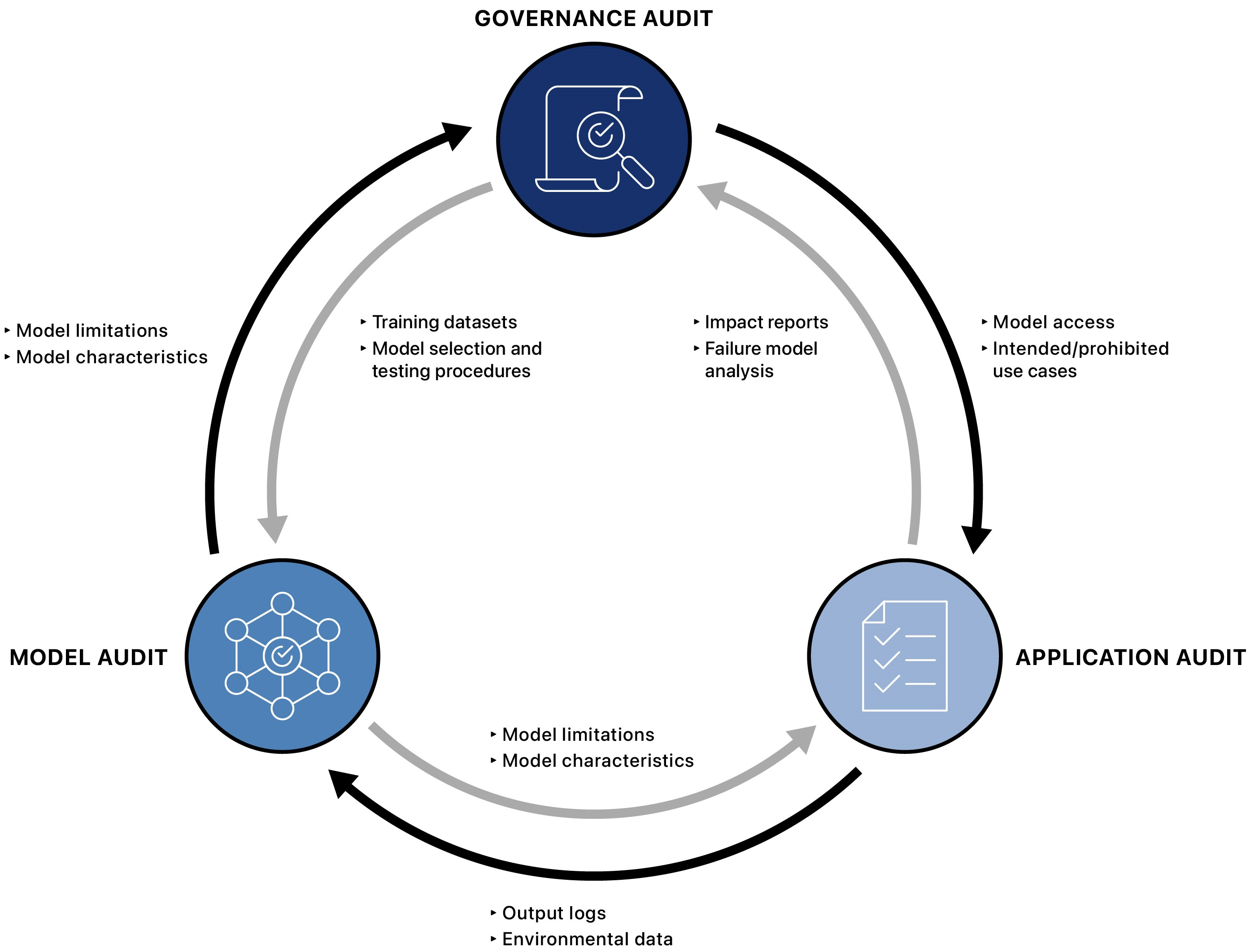}
    \caption{Outputs from audits on one level become inputs for audits on other levels.}
    \label{fig:2}
\end{figure}

Governance, model, and application audits must be connected into a structured process, meaning that outputs from one level become inputs on other levels (see \cref{fig:2}). Model audits, for instance, produce reports summarising LLMs' properties and limitations, which should inform application audits that verify whether a model's known limitations have been considered when designing downstream applications. Similarly, ex-post application audits produce output logs documenting the impact that different applications have in applied settings. Such logs should inform LLMs' continuous redesign and revisions of their accompanying model cards. Governance audits must check the extent to which technology providers' software development processes include mechanisms to incorporate feedback from application audits.

Each step in our three-layered approach should involve independent third-party auditors (\hyperlink{claim2}{Claim 2}). However, two caveats are required here. First, it need not be the same organisation conducting audits on all three levels as each requires different competencies. Governance audits require understanding corporate governance \cite{cihonCorporate2021} and soft skills like stakeholder communication. Model audits are highly technical and require knowledge about evaluating ML models, operationalising different normative dimensions, and visualising model characteristics. Application auditors typically need domain-specific expertise. All these competencies may not be found within one organisation.

Second, as institutional arrangements vary between jurisdictions and sectors, the best option may be to leverage the capabilities of institutions operating within a specific geography or industry to perform various elements of governance, model, and application audits. For example, medical devices are already subject to various testing and certification procedures before being launched. Hence, application audits for new medical devices incorporating LLMs could be integrated with such procedures. In part, this is already happening. The US Food and Drug Administration (FDA) has proposed a regulatory framework for modifying ML-based software as a medical device \cite{fdaArtificial2021}. The core point is that different independent auditors can perform the three different types of audits outlined here and that different institutional arrangements may be preferable in different jurisdictions or sectors.

%% file: sections/5_limitations.tex
\section{Limitations and avenues for further research}
\label{sec:5}
This section highlights three limitations of our work that apply to any attempt to audit LLMs: one conceptual, one institutional and one practical. First, model audits pose conceptual problems related to constructing validity. Second, an institutional ecosystem to support independent third-party audits has yet to emerge. Third, not all LLM-related social and ethical risks can be practically addressed on the technology level. We consider these limitations in turn, discuss potential solutions, and provide directions for future research.

\subsection{Lack of methods and metrics to operationalise normative concepts}
One bottleneck to developing effective auditing procedures is the difficulty of operationalising normative concepts like robustness and truthfulness \cite{Jacobs2021}. A recent case study found that organisations' lack of standardised evaluation metrics is a crucial challenge when implementing AI auditing procedures \cite{mokanderOperationalising2022, mokanderChallenges2022}. The problem is rooted in construct validity, i.e., the extent to which a given metric accurately measures what it is supposed to \cite{smithResearch2014}. Construct validity problems primarily arise in our blueprint from attempts to operationalise characteristics like performance, robustness, information security and truthfulness during model audits.

Consider truthfulness as an example. LLMs do not require a model of the real world. Instead, they compress vast numbers of conditional probabilities by picking up on language regularities \cite{sobieszekPlaying2022}. Therefore, they have no reason to favour any reality but can select from various possible worlds, provided each is internally coherent \cite{reynoldsPrompt2021}.\footnote{LLMs favour the statistically most likely reality given their training data. However, any training data necessarily constitute a reduction of reality that supports some interpretations but obscures others \cite{smithPromise2019}.} However, different epistemological positions disagree about the extent to which this way of sensemaking is unique to LLMs or, indeed, a problem at all. Simplifying to the extreme, realists believe in objectivity and the singularity of truth, at least insofar as the natural world is concerned \cite{hackingRepresenting1983}. Whereas relativists believe that truth and falsity are products of context-dependent conventions and assessment frameworks \cite{rortyPragmatism2021}. Numerous compromise positions can be found on the spectrum between those poles. However, tackling pressing social issues cannot await the resolution of long-standing philosophical disagreements. Indeed, courts settle disagreements daily based on pragmatist operationalisations of concepts like truth and falsehood in keeping with the pragmatic maxim that theories should be judged by their success when applied practically to real-world situations \cite{leggPragmatism2020}.

Following that reasoning, we argue that refining pragmatist operationalisations of concepts like truthfulness and robustness do more to promote fairness, accountability, and transparency in using LLM than either dogmatic or sceptical alternatives. However, developing metrics to capture the essence of thick normative concepts is difficult and entails many well-known pitfalls. Reductionist representations of normative concepts generally bear little resemblance to real-life considerations, which tend to be highly contextual \cite{leeFormalising2022}. Moreover, different operationalisations of the same normative concept (like `fairness') cannot be satisfied simultaneously \cite{friedlerIm2021}. Finally, the quantification of normative concepts can itself have subversive or undesired consequences \cite{islamMetrics2021, cuguero-escofetEthics2017}. As Goodhart's Law reminds us, a measure ceases to be a good metric once it becomes a target.

The operationalisation of characteristics like performance, robustness, information security and truthfulness discussed in \cref{sec:4} is subject to the above limitations. Resolving all construct validity problems may be impossible, but some ways of operationalising normative concepts are better than others for evaluating an LLM's characteristics. Consequently, an important avenue for further research is developing new methods to operationalise normative concepts in ways that are verifiable and maintain high construct validity.

\subsection{Lack of an institutional ecosystem}
A further limitation is that our blueprint does not decisively identify who should conduct the three audits it recommends. This is a limitation, since any auditing procedure will only be as good as the institution delivering it \cite{boddingtonCode2017}. However, we have left the question open for two reasons. First, different institutional ecosystems intended to support audits and conformity assessments of AI systems are currently emerging in different jurisdictions and sectors \cite{minkkinenEcosystems2021}. Second, our blueprint is flexible enough to be adopted by any external auditor. Hence, the feasibility and effectiveness of our approach do not hinge on the question of institutional design.

That said, the question of who audits whom is important, and much can be learned from auditing in other domains. Five institutional arrangements for structuring independent audits are particularly relevant to our purposes. Audits of LLMs can be conducted by:
\begin{enumerate}
    \item \textit{Private service providers}, chosen by and paid for by the technology provider (equivalent to the role accounting firms play during financial audits or business ethics audits \cite{schopplEthics2022}).
    \item \textit{A government agency}, centrally administered and paid for by government, industry, or a combination of both (equivalent to the FDA's role in approving food and drug substances \cite{fdaInspection2022}).\footnote{Alternative models of government involvement exist. For example, audits may be conducted or sanctioned by a government agency like the National Institute of Standards and Technology (NIST) in the US or by the same notified bodies that the European Commission has tasked with performing conformity assessments of high-risk AI systems in the EU.}
    \item \textit{An industry body}, operationally independent yet funded through fees from its member companies (equivalent to the British Safety Council's role in audits of workers' health and safety \cite{councilBritish2023}).
    \item \textit{Non-profit organisations}, operationally independent and funded through public grants and voluntary donations (equivalent to the Rainforest Alliance role in auditing forestry practices \cite{allianceOur2023}).
    \item \textit{An international organisation}, administered and funded by its member countries (equivalent to the International Atomic Energy Agency's role in auditing nuclear medicine practices \cite{iaeaQuality2015}).
\end{enumerate}

Each of these arrangements has its own set of affordances and constraints. Private service providers, for example, are under constant pressure to innovate, which can be beneficial given the fast-moving nature of LLM research. However, private providers' reliance on good relationships with technology providers to remain in business increases the risk of collusion \cite{dufloTruthtelling2013}. Therefore, some researchers have called for more government involvement, including an ‘FDA for algorithms' \cite{tuttFDA2017}. Establishing a government agency to review and approve high-risk AI systems could ensure the uniformity and independence of pre-market audits but might stifle innovation and cause longer lead times. Moreover, while the FDA enjoys a solid international reputation \cite{carpenterReputation2014}, not all jurisdictions would consider the judgement of an agency with a national or regional mandate legitimate.

The lack of an institutional ecosystem to implement and enforce the LLM auditing blueprint outlined in this article is a limitation. Without clear institutional arrangements, claims that an AI system has been audited are difficult to verify and may exacerbate harms \cite{costanza-chockWho2022}. Further research could usefully investigate the feasibility and effectiveness of different institutional arrangements for conducting and enforcing the three types of audits proposed.

\subsection{Not all risks from LLMs can be addressed on the technology level}
Our blueprint for auditing LLMs has been designed to contribute to good governance. However, it cannot eliminate the risks associated with LLMs for at least three reasons. First, most risks cannot be reduced to zero \cite{nistRisk2002}. Hence, the question is not whether residual risks exist but how severe and socially acceptable they are \cite{fraserWhere2021}. Second, some risks stem from deliberate misuse, creating an offensive-defensive asymmetry wherein responsible actors constantly need to guard against all possible vulnerabilities while malicious agents can cause harm by exploiting a single vulnerability \cite{vanmerwijkAI2022}. Third, as we will expand on below, not all risks associated with LLMs can be addressed on the technology level. 

Weidinger et al. \cite{weidingerEthical2021b} list over 20 risks associated with LLMs divided into six broad risk areas. In \cref{sec:4}, we highlighted how our three-layered approach helps identify and mitigate some of these risks. To recap, amongst others, governance audits can help protect against risks associated with malicious use; model audits can help identify and manage information and misinformation hazards; and application audits can help protect against discrimination as well as experiential harms. Of course, these are just examples. Audits at each level contribute, directly or indirectly, to addressing many different risks. However, not all the risks listed by Weidinger et al. are captured by our blueprint. Consider automation harm as an example. Increasing the capabilities of LLMs to complete tasks that would otherwise require human intelligence threatens to undermine creative economies \cite{sautoyCreativity2019}. While some highly potent LLMs may remove the basis for some professions that employ many people today -- such as translators or copywriters -- that is not necessarily a failure of the technology. The alternative of building less capable LLMs is counterproductive since abstaining from technology usage generates significant social and economic opportunity costs \cite{floridiAI4People2018}.

The problem is not necessarily change per se but its speed and how the fruits of automation are distributed \cite{freyTechnology2019, mokanderAlgorithmic2022}. Hence, problems related to changing economic environments may be better addressed through social and political reform rather than audits of specific technologies. It is important to remain realistic about auditing's capabilities and not fall into the trap of overpromising when introducing new governance mechanisms \cite{sloaneAlgorithmic2021}. However, the fact that no auditing procedures can address all risks associated with LLMs does not diminish their merits. Instead, it points towards another important avenue for further research: how can and should social and political reform complement technically oriented mechanisms in holistic efforts to govern LLMs?

%% file: sections/6_conclusion.tex
\section{Conclusion}
\label{sec:6}
Some of the features that make LLMs attractive also create significant governance challenges. For instance, the potential to adapt LLMs to a wide range of downstream applications undermines system verification procedures that presuppose well-defined demand specifications and predictable operating environments. Consequently, our analysis in \cref{sec:3} concluded that existing AI auditing procedures are not well-equipped to assess whether the checks and balances put in place by technology providers and downstream developers are sufficient to ensure good governance of LLMs.

In this article, we have attempted to bridge that gap by outlining a blueprint for auditing LLMs. In \cref{sec:4}, we introduced a three-layered approach, whereby governance, model and application audits inform and complement each other. During \textit{governance audits}, technology providers' accountability structures and quality management systems are evaluated for robustness, completeness, and adequacy. During \textit{model audits}, LLMs' capabilities and limitations are assessed along several dimensions, including performance, robustness, information security, and truthfulness. Finally, during \textit{application audits}, products and services built on top of LLMs are first assessed for legal compliance and subsequently evaluated based on their impact on users, groups, and the natural environment.

Technology providers and policymakers have already started experimenting with some of the auditing activities we propose. Consequently, auditors can leverage a wide range of existing tools and methods, such as impact assessments, benchmarking, model evaluation, and red teaming, to conduct governance, model, and application audits. That said, the feasibility and effectiveness of our three-layered approach hinge on two factors. First, only when conducted in a combined and coordinated fashion can governance, model and application audits enable different stakeholders to manage LLM-related risks. Hence, audits on the three levels must be connected in a structured process. Governance audits should ensure that providers have mechanisms to take the output logs generated during application audits into account when redesigning LLMs. Similarly, application audits should ensure that downstream developers take the limitations identified during model audits into account when building on top of a specific LLM. Second, audits must be conducted by an independent third-party to ensure that LLMs are ethical, legal, and technically robust. The case for independent audits rests not only on concerns about the misaligned incentives that technology providers may face but also on concerns about the rapidly increasing capabilities of LLMs \cite{zieglerAdversarial2022}.

However, even when implemented under ideal circumstances, audits will not solve all tensions or protect against all risks of harm associated with LLMs. So, it is important to remain realistic about what auditing can achieve. Three limitations of our approach are worth reiterating. First, the feasibility of the model audits hinges on the construct validity of the metrics used to assess model characteristics like robustness and truthfulness. This is a limitation because it is notoriously difficult to operationalise normative concepts. Second, our blueprint for auditing LLMs does not specify who should conduct the audits it posits. No auditing procedure is stronger than the institutions backing it. Hence, the fact that an ecosystem of actors capable of implementing our blueprint has yet to emerge constrains its effectiveness. Third, not all risks associated with LLMs arise from processes that can be addressed through auditing. Some tensions are inherently political and require continuous management through public deliberation and structural reform.

Academics and industry researchers can contribute to overcoming these limitations by focusing on two avenues for further research. The first is to develop new methods and metrics to operationalise normative concepts in ways that are verifiable and maintain a high degree of construct validity. The second is to disentangle further the sources of different types of risks associated with LLMs. Such research would advance our understanding of how political reform can complement technically oriented mechanisms in holistic efforts to govern LLMs.

Policymakers can facilitate the emergence of an institutional ecosystem capable of carrying out and enforcing governance, model, and application audits of LLMs. For example, policymakers can encourage and strengthen private sector LLM auditing initiatives by supporting the standardisation of evaluation metrics \cite{keyesMulching2019}, harmonising AI regulation \cite{mokanderUS2022}, facilitating knowledge sharing \cite{epsteinTuringbox2018} or rewarding achievements through monetary incentives \cite{floridiAI4People2018}. Policymakers should also update existing and proposed AI regulations in line with our three-layered approach to address LLM-related risks. For example, while the EU AI Act's conformity assessments and post-market monitoring plans mirror application audits, the proposed regulation does not contain mechanisms akin to governance and model audits \cite{mokanderConformity2022}. Without amendments, such regulations are unlikely to generate adequate safeguards against the risks associated with LLMs.

Our findings most directly concern technology providers as they are primarily responsible for ensuring that LLMs are legal, ethical, and technically robust. Such providers have moral and material reasons to subject themselves to independent audits, including the need to manage financial and legal risks \cite{eprsGovernance2019} and build an attractive brand \cite{eiuStaying2020}. So, what ought technology providers do? Firstly, they should subject themselves to governance audits and their LLMs to model audits. That would create a demand for independent auditing and accreditation bodies and help spark methodological innovation in governance and model audits. Secondly, providers should demand that products and services built on top of their LLMs undergo application audits. That could be done through structured access procedures, whereby permission for using an LLM is conditional on such terms. Thirdly, like-minded providers should establish, and fund, an independent industry body that conducts or commissions governance, model, and application audits.

Taking a long-term perspective, our three-layered approach holds lessons for how to audit more capable and general future AI systems. This article has focused on LLMs because they have broad societal impacts via widespread applications. However, elements of the governance challenges – including generativity, emergence, lack of grounding, and lack of access – have some general applicability to other ML-based systems \cite{Mondal2023, Muller2022}. Hence, we anticipate that our blueprint can inform the design procedures for auditing other generative, ML-based technologies.

That said, the long-term feasibility and effectiveness of our blueprint for how to audit LLMs may also be undermined by future developments. For example, governance audits make sense when only a limited number of actors have the ability and resources to train and disseminate LLMs. However, the democratisation of AI capabilities – either through the reduction of entry barriers or a turn to business models based on open-source software – would challenge this status quo \cite{Rao2020}. Similarly, if language models become more fragmented or personalised \cite{Kirk2023}, there will be many user-specific branches or instantiations of a single LLM which would make model audits more complex to standardise. As a result, while maintaining the usefulness of our three-layered approach, we acknowledge that it will need to be continuously revised in response to the changing technological and regulatory landscape.

It is worth concluding with some words of caution. Our blueprint is not intended to replace existing governance mechanisms but to complement and interlink them by strengthening procedural transparency and regularity. Rather than being adopted wholesale by technology providers and policymakers, we hope that our three-layered approach can be adopted, adjusted, and expanded to meet the governance needs of different stakeholders and contexts.

%% file: sections/7_appendix.tex
\newpage
\appendix
\section{Methodology}
\label{sec:appendix}
Before describing our methodology, something should be said about our research approach. According to the pragmatist tradition, research is only legitimate when applied, i.e., grounded in real-world problems \cite{salkindEncyclopedia2010}. As established in \cref{sec:2}, there is a need to develop new governance mechanisms that different stakeholders can use to identify and mitigate the risks associated with LLMs. In this article, we take a pragmatist stance when exploring how auditing procedures can be designed so they are feasible and effective in practice.

Designing procedures to audit LLMs is an art, not a science. In a policy context, applied research concerns the evaluation of different governance design decisions or policies in relation to a desired outcome \cite{haasApplied1998}. From a pragmatist point of view, however, a mark of quality in applied policy research is that questions are answered in ways that are actionable \cite{leggPragmatism2020}. That implies that researchers must sometimes go beyond an evaluation of existing options to prescribe new solutions. While there is no guarantee that the best course of action will be found, researchers can ensure rigour by systematically building on previous research and by incorporating input from different stakeholders.

Mindful of those considerations, the following methodology was used to develop our blueprint for how to audit LLMs. Note that while the five steps below exhaust the range of research activities that went into this study, the sequential presentation is a gross simplification. In reality, the research process was messy and iterative, with several of the steps overlapping both thematically and chronologically.

Firstly, we mapped existing auditing procedures designed to identify the risks associated with different AI systems through a systematised literature review \cite{grantTypology2009}. In doing so, we searched five databases (Google Scholar, Scopus, SSRN, Web of Science and arXiv) for articles related to the auditing of AI systems. Keywords for the search included (``auditing'', ``evaluation'' OR ``assessment'') AND (``fairness'', ``truthfulness'', ``transparency'' OR ``robustness'') AND (``language models'', ``artificial intelligence'' OR ``algorithms''). However, not all relevant auditing procedures have been developed by academic researchers. Hence, we used a snowballing technique \cite{wohlinGuidelines2014}, i.e., tracking the citations of already included articles, to identify auditing procedures developed by private service providers, national or regional policymakers and industry associations. A total of 126 documents were included in this systematised literature review.

Secondly, we identified the elements underpinning these procedures. This resulted in a typology that distinguishes between different types of audits, e.g., risk and compliance audits; internal and external audits; ex-ante and ex-post audits; as well as between functionality, code, and impact audits. The space of possible auditing procedures consists of all unique combinations between these different elements.

Thirdly, we generated a list of key claims about how auditing procedures for LLMs should be designed so that they are feasible and effective in practice. To do this, we conducted a gap analysis between the governance challenges posed by LLMs on the one hand and the theoretical affordances of existing AI auditing procedures on the other. Our analysis resulted in six key claims about how auditing procedures should be designed in order to capture the full range of risks posed by LLMs. Those claims are presented and discussed in \cref{sec:3}.

Fourthly, we created a draft blueprint for how to audit LLMs by identifying the smallest set of auditing procedures that satisfied our six key claims. In practice, not all auditing procedures are equally effective in identifying the risks posed by LLMs. Besides, some auditing procedures serve similar functions. Although some redundancy is an important feature in safety engineering, too much overlap between different auditing regimes can be counterproductive in so far as roles and responsibilities become less clear and scarce resources are being consumed that could otherwise have been more effectively invested elsewhere. This step thus consisted of reducing the theoretical space of possible auditing procedures into a limited set of activities that are (1) jointly sufficient to identify the full range of risks associated with LLMs, (2) practically feasible to implement, and (3) seem to have a justifiable cost-benefit ratio.

Fifthly and finally, we sought to refine and validate our draft blueprint by triangulating findings \cite{freySAGE2018} from different sources. For example, we sought input from a diverse set of stakeholders. In total, we conducted over 20 semi-structured interviews \cite{adamsConducting2015} with, and received feedback from, researchers, professional auditors, AI developers at frontier labs and policymakers in different jurisdictions. The final blueprint outlined in \cref{sec:4} is the result of those consultations.